\pdfoutput=1

\documentclass[11pt]{article}

\usepackage[final]{acl}

\usepackage{times}
\usepackage{latexsym}

\usepackage[T1]{fontenc}

\usepackage[utf8]{inputenc}

\usepackage{microtype}

\usepackage{inconsolata}

\usepackage{graphicx}

\title{KARPA: A Training-free Method of Adapting Knowledge Graph as References for Large Language Model's Reasoning Path Aggregation}

\author{Siyuan Fang$^1$, Kaijing Ma$^2$, Tianyu Zheng$^3$, Xinrun Du$^3$,\\
\textbf{Ningxuan Lu$^4$, Ge Zhang$^3$, Qingkun Tang$^5$\thanks{Corresponding Author.}}
    \\[3pt]
    $^1$Beijing University of Posts and Telecommunications, $^2$Tongji University,\\ $^3$Multimodal Art Projection Research Community, $^4$Duke University,
    $^{5}$ZTE Corporation\\[1pt]
\small \texttt{syfang@bupt.edu.cn, tang.qingkun@zte.com.cn} \\[2pt]
\vspace{-4ex}
}

\begin{document}
\maketitle
\begin{abstract}
Large language models (LLMs) demonstrate exceptional performance across a variety of tasks, yet they are often affected by hallucinations and the timeliness of knowledge. Leveraging knowledge graphs (KGs) as external knowledge sources has emerged as a viable solution, but existing methods for LLM-based knowledge graph question answering (KGQA) are often limited by step-by-step decision-making on KGs, restricting the global planning and reasoning capabilities of LLMs, or they require fine-tuning or pre-training on specific KGs. To address these challenges, we propose \textbf{Knowledge graph Assisted Reasoning Path Aggregation (KARPA)}, a novel framework that harnesses the global planning abilities of LLMs for efficient and accurate KG reasoning. KARPA operates in three steps: pre-planning relation paths using the LLM’s global planning capabilities, matching semantically relevant paths via an embedding model, and reasoning over these paths to generate answers. Unlike existing KGQA methods, KARPA avoids stepwise traversal, requires no additional training, and is adaptable to various LLM architectures. Extensive experimental results show that KARPA achieves state-of-the-art performance in KGQA tasks, delivering both high efficiency and accuracy. Our code will be available on Github.
\end{abstract}

\section{Introduction}

In recent years, large language models (LLMs) \citep{touvron2023llama,touvron2023llama2,achiam2023gpt,qwen} have revolutionized natural language processing, demonstrating impressive performance in areas such as information extraction \citep{xu2023large}, summarization \citep{jin2024comprehensive}, and question answering \citep{louis2024interpretable}. However, despite these advancements, LLMs face notable challenges, particularly in maintaining up-to-date knowledge, domain-specific knowledge \citep{zhang2024knowledge}, and dealing with hallucinations \citep{zhang2023hallucination,huang2023survey} where LLMs produce incorrect or nonsensical outputs.

\begin{figure*}[t]
\centering
\includegraphics[width=0.85\textwidth]{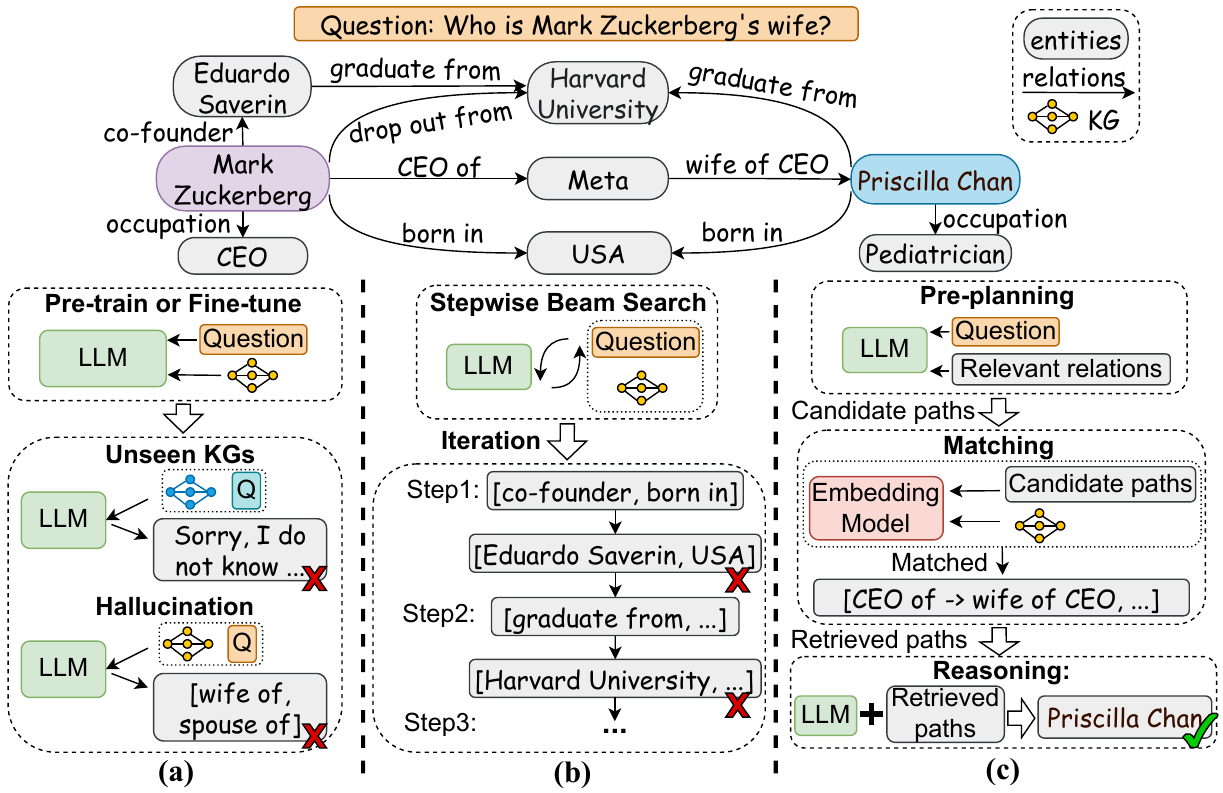} 
\caption{Comparison of different LLM-based KGQA methods: (a) Pre-training or fine-tuning the LLM on specific KG data. (b) Direct reasoning over KGs via stepwise interactions with the LLM. (c) Our KARPA framework, which combines the global planning and reasoning abilities of LLMs with embedding-based matching techniques. KARPA enables comprehensive path planning while avoiding local optima and hallucinations observed in previous methods.}
\label{figure1}
\end{figure*}

Knowledge graphs (KGs) enhance the reasoning capabilities of LLMs by providing structured, reliable external knowledge \citep{zhu2024llms,pan2024unifying}. Existing approaches to integrating LLMs with KGs fall into two categories: (1) Direct interaction between LLMs and KGs, where the LLM explores the KG step-by-step \citep{sun2023think,jiang2023structgpt}, often relying on local search strategies like beam search. These methods can produce suboptimal answers by overlooking the LLM's global planning and reasoning potential. Additionally, they require numerous interactions between LLMs and KGs, as shown in Figure~\ref{figure1}(b). (2) Training-based methods, such as reasoning on graphs (RoG) \citep{luo2023reasoning}, generate retrieval information for KGQA. However, they often require fine-tuning or pre-training on specific KG data \citep{li2023trea,huang2024joint}. These methods struggle with unseen KGs, necessitate retraining, and are prone to hallucinations during information generation, as illustrated in Figure~\ref{figure1}(a).

To address these limitations, we propose \textbf{Knowledge graph Assisted Reasoning Path Aggregation (KARPA)}, an innovative framework that leverages the global planning capabilities of LLMs alongside semantic embedding models for efficient and accurate KG reasoning. Our approach consists of three key steps: pre-planning, matching, and reasoning, as shown in Figure~\ref{figure2}. In the pre-planning phase, KARPA enables the LLM to generate initial relation paths for the provided question using LLM's inherent reasoning and planning capabilities. With these inital relation paths, KARPA employs a semantic embedding model \citep{ruder2019survey} to identify candidate relations that are semantically similar to the relations within the initial paths. The LLM can then create coherent relation paths that logically connect the topic entity to potential answer entities using these candidate relations. During the matching phase, KARPA employs an embedding model to identify relation paths within the KG that exhibit the highest similarity to the paths generated by the LLM in the pre-planning phase. This avoids locally optimal issues encountered in previous methods. Finally, during the reasoning step, the matched relation paths and their corresponding entities are provided to the LLM to formulate final answers. The detail of our framework is shown in Figure~\ref{figure2}.

KARPA offers several key advantages over existing LLM-based KGQA methods: (1) KARPA fully leverages the global planning and reasoning capabilities of LLMs to generate logically coherent paths, which aligns better with human-like reasoning processes. Unlike methods limited to adjacent relations or requiring iterative traversal within the KG, KARPA selects from all potential relations within the KG, significantly reducing interactions between LLMs and KGs. (2) Our embedding-based matching strategy avoids the locally optimal solution that arises from the stepwise interactions between LLMs and KGs, ensuring more effective exploration of the KGs. (3) KARPA is training-free, making it adaptable to various LLMs while enhancing reasoning capabilities with techniques such as chain-of-thought (CoT) \citep{wei2022chain}. Our contributions can be summarized as follows:

\begin{itemize}
    \item We propose KARPA, which combines the global planning and reasoning capabilities of LLMs with embedding models to improve both accuracy and efficiency of KGQA tasks.
    \item By enabling LLMs to generate initial relation paths across all potential relations within the KG and integrating a semantic embedding model for path matching, KARPA mitigates the risk of local optima and minimizes interactions with KGs. Techniques such as CoT can also be incorporated to further enhance the LLM's reasoning abilities over KGs.
    \item KARPA operates in a training-free manner and is compatible with various LLMs, providing a plug-and-play solution that achieves state-of-the-art performance on several KGQA benchmark datasets.

\end{itemize}

\section{Related Work}

\paragraph{Prompt-Based Reasoning with LLMs.} LLMs such as LLaMA \citep{touvron2023llama,touvron2023llama2}, Qwen \citep{qwen}, and GPT-4 \citep{achiam2023gpt}have advanced reasoning by leveraging extensive internal knowledge. Various prompt-based methods further enhance these capabilities. For instance, Chain-of-Thought (CoT) \citep{wei2022chain} improves reasoning by decomposing complex tasks into manageable steps, excelling in domains like mathematical reasoning \citep{jie2023design} and logical inference \citep{zhao2023enhancing}. Variants such as Auto-CoT \citep{zhang2022automatic}, Zero-Shot-CoT \citep{kojima2022large}, and Complex-CoT \citep{fu2022complexity} further optimize this approach. Frameworks like Tree of Thoughts (ToT) \citep{yao2024tree} and Graph of Thoughts (GoT) \citep{besta2024graph} have expanded the scope of LLM reasoning. Lately, OpenAI o1 series models represent a significant advancement in LLM reasoning. These methods underscore the role of tailored prompts in maximizing LLM reasoning potential.

\paragraph{LLM-Based KGQA.} Integrating KGs with LLMs enhances reasoning and mitigates hallucinations. Unlike methods such as CoT that rely solely on the internal knowledge of LLMs, incorporating KGs provides access to structured external knowledge \citep{he2022rethinking,wang2023knowledge}. Approaches like Think-on-Graph (ToG) \citep{sun2023think}, Interactive-KBQA \citep{xiong2024interactive} and StructGPT \citep{jiang2023structgpt} enable stepwise interactions between LLMs and KGs. Methods such as Reasoning on Graphs (RoG) \citep{luo2023reasoning}, chain of knowledge \citep{li2023chain} and other techniques \citep{huang2024joint,pan2024unifying,li2023trea} utilize pre-trained or fine-tuned LLMs to generate retrieval information for KGQA. Furthermore, methods like UniKGQA \citep{jiang2022unikgqa} and KG-CoT \citep{zhaokg} require training specific models for KG information retrieval, further complicating their implementation.

\section{Preliminary}

\textbf{Knowledge Graphs (KGs).} KGs represent structured information as \( G = (E, R) \), where \( E \) is the set of entities and \( R \) denotes the set of relations. Each relation \( r \in R \) connects two entities \( (e_i, e_j) \), with \( e_i, e_j \in E \).

\textbf{Relation Paths and Reasoning Paths.} A relation path \( P \) connects a topic entity \( e_t \) to an answer entity \( e_a \) via a sequence of relations: \( P = (r_1, r_2, \ldots, r_n) \), where \( r_i \in R \). Reasoning paths further include intermediate entities along the path, represented as \( P_r = \left\{e_t \overset{r_1}{\rightarrow} e_1 \overset{r_2}{\rightarrow} \ldots \overset{r_n}{\rightarrow} e_a \right\} \).

\textbf{Knowledge Graph Question Answering (KGQA).} KGQA aims to answer questions using information from KGs. Given a query \( Q \), the goal of KGQA is to generate an answer \( A \) using a function \( f \): \( A = f(Q, G) \), where \( f \) extracts the answer from the KG \( G \) based on \( Q \).

\textbf{Embedding Models and Semantic Similarity.} Embedding models represent text in a continuous vector space, enabling semantic similarity measurements. A function \( \Phi: R \to \mathbb{R}^d \) maps a sentence \( R \) to a \( d \)-dimensional vector. Similarity between embeddings is computed using cosine similarity:
\begin{equation}
    sim(r_i, r_j) = \frac{\Phi(r_i) \cdot \Phi(r_j)}{\|\Phi(r_i)\| \|\Phi(r_j)\|},
    \label{eq1}
\end{equation}
where \( \cdot \) is the dot product and \( \|\cdot\| \) is the Euclidean norm. This metric aids in comparing semantic information for retrieval tasks.

\begin{figure*}[t]
\centering
\includegraphics[width=0.98\textwidth]{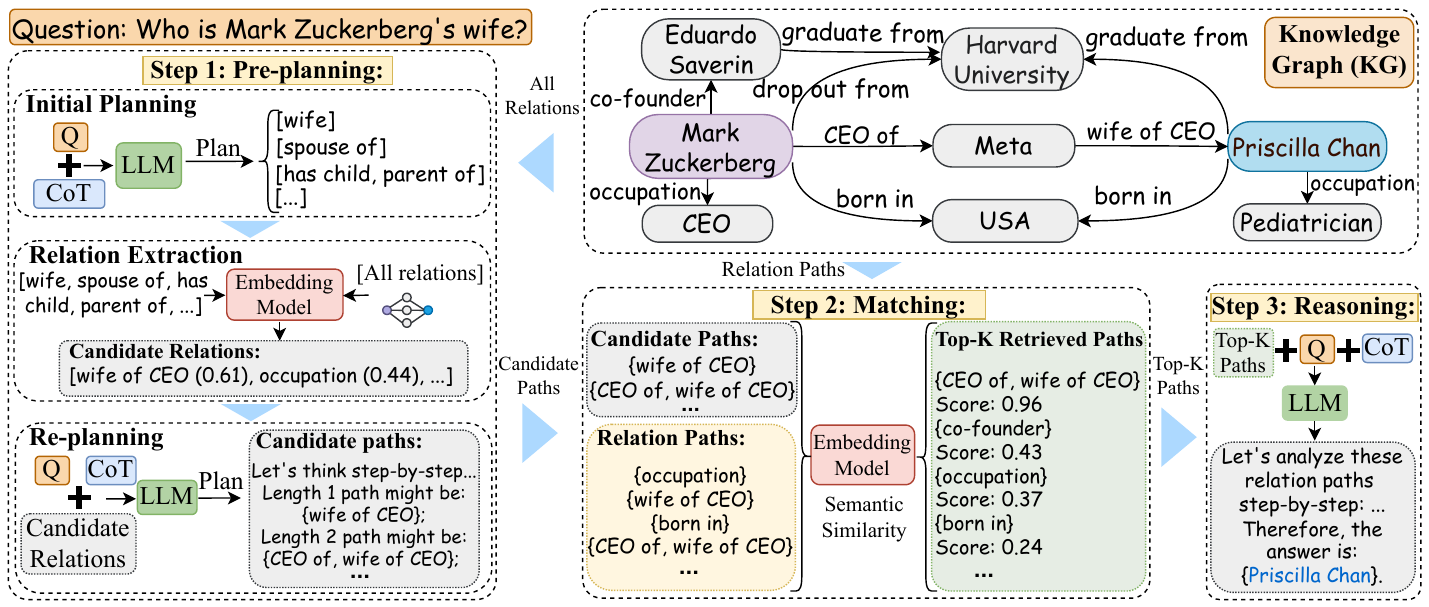}
\caption{The framework of our KARPA. Our framework consists of three main steps: (1) Pre-planning: The LLM generates initial relation paths based on the given question, decomposes them for relation extraction, and re-plans coherent candidate paths that connect the topic and answer entities with relevant relations. (2) Matching: Relation paths are extracted based on their similarity to the re-planned candidate paths using an embedding model. Our matching method accommodates paths of varying lengths. (3) Reasoning: The selected top-$K$ paths, combined with the question and corresponding entities, form a prompt for the LLM to enable accurate KG question answering.}
\label{figure2}
\end{figure*} 

\section{Approach}

In this section, we present KARPA, a framework that leverages the strengths of LLMs and embedding models to enhance KGQA. Our approach is composed of three key steps: pre-planning, matching, and reasoning, as illustrated in Figure~\ref{figure2}.

\subsection{Pre-Planning with LLM}

The pre-planning phase leverages the global planning capabilities of LLMs to generate initial paths \(P_{initial}\) and candidate paths \(P_{cand}\). This phase initiates the reasoning process by allowing the LLM to analyze the input question \(Q\) and the associated topic entity \(e_{t}\). By leveraging the reasoning capability of LLM, KARPA is able to propose paths that are not only logically coherent but also have the potential to lead to the answer entities \(E_{a}\).

\paragraph{Initial Planning Using LLM} KARPA starts by using the LLM to generate a set of initial relation paths based on the provided question \(Q\), as shown in Figure~\ref{figure2}. The LLM outputs a set of potential relation paths \(P\) as follows:
\begin{equation}
\begin{split}
    P = \{p_1, p_2, \ldots, p_m\}, \\ 
    \text{where } p_i = (r_1^i, r_2^i, \ldots, r_{n_i}^i).
    \label{eq2}
\end{split}
\end{equation}
Here, each \(p_i\) is a path of \(n_i\) relations \(r_j^i\) that could logically connect a topic entity \(e_{t}\) to the potential answer entity \(e_{a}\). The relations within these paths serve as candidates for relations extraction.

\paragraph{Relation Extraction Strategy}

With the initial relation paths \(P\), we decompose each path \(p_i\) into its constituent relations $R_i = \{r_1^i, r_2^i, \ldots, r_{n_i}^i\}$. For each relation \(r_j^i \in R_i\), we utilize an embedding model to identify top-\(K\) semantically similar relations from the entire KG:
\begin{equation}
    R_j^i = \{r_{j1}, r_{j2}, \ldots, r_{jk}\} = \text{Top-K}(\text{sim}(\mathbf{r_j^i}, \mathbf{r})),
    \label{eq3}
\end{equation}
where \(\text{sim}(\cdot)\) calculates the semantic similarity function (e.g., cosine similarity) between the embedding of relation \(r_j^i\) and all relations \(r \in KG\) using Equation~\ref{eq1}. The resulting set \(R_j^i\) contains the relations that best align with the initial relations, ensuring that LLM has access to all relevant relations beyond the immediate neighbors in the KG.

\paragraph{Re-planning Relation Paths with LLM}

Using the candidate relations \(R_j^i\) from the previous step, the LLM constructs refined relation paths \(P_{cand}\) that potentially connect the topic entity \(e_{t}\) to the answer entity \(e_{a}\):
\begin{equation}
    P_{cand} = \text{LLM}(Q, R_j^i),\ \text{each } r_j^i \in  R_j^i \subset R.
    \label{eq4}
\end{equation}
Given the question \(Q\) and candidate relations \(R_j^i\), the LLM utilizes its reasoning capabilities to produce coherent candidate paths \(P_{cand}\), as shown in Figure~\ref{figure2}. During this phase, techniques like CoT can be incorporated to strengthen LLM's logical reasoning, ensuring the construction of semantically meaningful paths.

By extracting relations from the entire KG rather than limiting to adjacent neighbors, KARPA avoids stepwise interactions, reducing the risk of local optima and unnecessary interactions with the KG. The pre-planning phase sets the foundation for efficient and accurate matching and reasoning in the subsequent steps.

\subsection{Relation Paths Matching}

The matching step in KARPA extracts relevant relation paths from KGs based on the LLM-generated candidate paths \(P_{cand}\), as shown in Figure~\ref{figure2}. This process systematically explores and scores potential relation paths for reasoning step.

\subsubsection{Conventional Relation Paths Matching}

Conventional LLM-based KG exploration methods, such as ToG\citep{sun2023think}, typically involve the LLM selecting top-\(K\) promising relations \(R_t\) from the adjacent relations of the current entity \(e\) at each step. This strategy resembles greedy algorithms, such as beam search. Formally, let \(R(e)\) denote the set of relations available for the current entity \(e\). The selection process can be defined as:
\begin{equation}
    R_{\text{selected}} = \text{argmax}_{r \in R(e)} \, f(r),\ r \in KG.
    \label{eq5}
\end{equation}
In Equation~\ref{eq5}, \(f(r)\) is a scoring function indicating the potential of relation \(r\). Since embedding similarity represents the similarity between two relations, we use $1-sim(r_i, r_j)$ as the cost function for beam search. However, this approach does not guarantee finding the optimal path, as it may overlook globally optimal solutions.

To enhance relation path matching, we employ traditional pathfinding algorithms like Dijkstra’s, which can be expressed as: 
\begin{equation}
    cost(v) = \min \{cost(v), cost(v') + cost(v', v)\}.
    \label{eq6}
\end{equation}
In Equation~\ref{eq6}, the cost to reach node \(v\) is determined by either its current known cost or the cost of reaching one of its predecessors \(v'\) plus \(cost(v', v)\), the cost of the edge connecting \(v'\) to \(v\).

In KARPA, we begin from the topic entity \(e_t\) and compute the semantic similarity \(sim(r_i, r_j)\) using Equation~\ref{eq1} for relations at each step, scoring the relations based on their similarity to the corresponding relations in the candidate relation paths \(P_{cand}\). The cost for each step is defined as: $cost(r) = 1 - sim(r_i, r_j)$. This modification ensures that higher similarity scores correspond to lower costs, facilitating optimal path discovery. Since similarity scores range from 0 to 1, we average the total cost of relation paths of different lengths so that shorter paths can be fairly compared with longer paths. The path matching function based on Dijkstra's algorithm can be defined as:
\begin{equation}
\begin{split}
    & cost(e) = \min \Big\{ \frac{1}{n_e} cost(e), \\
    \frac{1}{n_{e'}+1}&\left[cost(e') + sim(r_{(e', e)},r_{cand})\right]\Big\},
    \label{eq7}
\end{split}
\end{equation}
where the cost of entity \(e\) is compared between \(cost(e)\) averaged by the number of relations \(n_e\) to reach entity \(e\), and the cost of its predecessor \(cost(e')\) plus the current cost $sim(r_{(e', e))},r_{cand})$, averaged by number of relations \(n_{e'}\) plus one.

\subsubsection{Heuristic Value-Based Paths Matching}

Since the conventional relation paths matching methods require the cost of each relations alone the paths, the similarity between initial relation paths and current paths within the KG can only be calculated when current paths have the same length as candidate paths \(P_{cand}\). Inspired by the heuristic value in A* algorithm, we design a heuristic value-based relation paths matching method. In the traditional A* algorithm, the heuristic value serves as a guiding function that indicates the distance between current node and target node. In KARPA, the heuristic value \(h\) indicate the semantic similarity between the candidate relation paths \(P_{cand}\) and current path within the KG. By using heuristic value \(h\) as an indicator, we are able to compute the similarity between paths of differing lengths, such as $A\xrightarrow[]{father}\xrightarrow[]{father}B$ and $A\xrightarrow[]{grandfather}B$, as shown in Figure~\ref{figure2}. For paths \(P_a\) and \(P_b\), we concatenate all relations into one sentence and use the embedding model to calculate their similarity:
\begin{equation}
    sim(P_a, P_b) = sim(\text{concat}(R_{Pa}), \text{concat}(R_{Pb})).
    \label{eq8}
\end{equation}
In Equation~\ref{eq8}, the similarity between path \(P_a\) and \(P_b\) can be calculated using the concatenation of their internal relations $R_P$ with Equation~\ref{eq1}. Since the heuristic value represents the semantic distance between \(P_a\) and \(P_b\), it can be defined as \(h = 1 - sim(P_a, P_b)\). The top-\(K\) relation paths \(P_K\) with lowest heuristic value can be extracted as:
\begin{equation}
    P_{K} = \text{argmax}_{P \in P_{all}} sim(P, P_{cand}),\ P_{all} \in KG.
    \label{eq9}
\end{equation}
Through Equation~\ref{eq9}, we are able to identify the top-\(K\) relevant paths from a diverse range of lengths as retrieved paths \(P_K\) for further reasoning.

The relation paths matching method in KARPA broadens the search space and mitigates the risk of missing potentially optimal paths that traditional methods might overlook. By dynamically adapting to paths of varying lengths, KARPA identifies top-\(K\) paths for LLM reasoning, ensuring robust and comprehensive path matching.

\subsection{Reasoning with LLM}

In the reasoning step, we combine the matched relation paths with their respective entities $e$ into a prompt for the LLM to reference during the final answer determination, as shown in Figure~\ref{figure2}. The reasoning process can be expressed as:
\begin{equation}
\begin{split}
    Answer = \text{LLM}(Q, P_K, e),\\
    P_K = \{r_1,r_2,\dots,r_n\}.
    \label{eq10}
\end{split}
\end{equation}
Given the top-\(K\) candidate paths \(P_K\) and their corresponding entities $e$, the LLM can effectively assess whether the provided connections lead to a valid answer to the question \(Q\). The KARPA framework facilitates the LLM's ability to evaluate multiple reasoning paths in parallel, thereby enhancing the overall efficiency of LLM-based KGQA tasks.

\begin{table*}[t]
\begin{center}
\resizebox{1.0\textwidth}{!}{
    \begin{tabular}{lccccccc}
        \toprule
        & & \multicolumn{3}{c}{WebQSP} & \multicolumn{3}{c}{CWQ} \\
        \cmidrule(lr){3-5} \cmidrule(lr){6-8}
        Type of Model & Method & Accuracy & Hit@1 & F1 & Accuracy & Hit@1 & F1 \\
        \midrule
        \multicolumn{8}{c}{Answering with Internal Knowledge} \\
        \midrule
        GPT-4 & IO prompt & - & 62.5 & - & - & 44.3 & - \\
        GPT-4 & CoT* \citep{sun2023think} & - & 67.3 & - & - & 46.0 & - \\
        \midrule
        \multicolumn{8}{c}{Training-based Methods} \\
        \midrule
        LLaMA2-7B (Fine-tune) & KD-CoT* \citep{wang2023knowledge} & - & 68.6 & 52.5 & - & 55.7 & - \\ 
        Graph Reasoning Model & KG-CoT* \citep{zhaokg} & - & \fbox{84.9} & - & - & 62.3 & - \\
        FiD-3B & DECAF* \citep{yu2022decaf} & - & 82.1 & \textbf{78.8} & - & \fbox{70.4} & - \\
        PLM (Pretrain)  & UniKGQA* \citep{jiang2022unikgqa} & - & 77.2 & \underline{72.2} & - & 51.2 & 49.0 \\
        LLaMA2-7B (Fine-tune) & RoG & \underline{80.4} & 84.6 & 70.1 & \fbox{60.5} & 61.3 & \fbox{54.2} \\
        \midrule
        \multicolumn{8}{c}{Direct Inference over KGs with LLMs} \\
        \midrule
        GPT-4o  & ToG & 58.6 & 78.5 & 50.9 & 53.3 & 56.8 & 41.9  \\
        GPT-4  & ToG* \citep{sun2023think} & - & 82.6 & - & - & 69.5 & - \\
        GPT-4  & Interactive-KBQA* & - & - & 71.2 & - & - & 49.1 \\
        GPT-4o  & KARPA & \fbox{76.1} & \underline{87.7} & 69.2 & \underline{69.8} & \underline{75.3} & \underline{58.4} \\
        GPT-4  & KARPA & \textbf{80.9} & \textbf{91.2} & \fbox{72.1} & \textbf{73.6} & \textbf{78.4} & \textbf{61.5} \\
        \bottomrule
\end{tabular}}
\caption{Performance comparison of KARPA with three method categories: (1) Answering with internal knowledge of LLMs, (2) Training-based methods, which require constant re-train for unseen KGs, and (3) Direct inference over KGs with LLMs. *Results are cited from corresponding publications. \textbf{Bold} represents the best result, \underline{underline} represents the second best, and \fbox{fbox} represents the third best.}
\vspace{-0.4cm}
\label{tab1}
\end{center}
\end{table*}

\begin{table}[h]
    \centering
    \resizebox{0.94\linewidth}{!}{%
    \begin{tabular}{@{}lcccc@{}}
        \toprule
        \textbf{Method} & \textbf{Dataset} & \textbf{Accuracy} & \textbf{Hit@1} & \textbf{F1} \\
        \midrule
        \multirow{3}{*}{RoG} 
        & Original & 63.5 & 77.8 & 64.8\\
        & Anonymized & 51.4 & 64.3 & 52.9\\
        & Variation & -12.1 & -13.5 & -11.9\\
        [2pt]\hdashline\\[-8pt]
        \multirow{3}{*}{ToG} 
        & Original & 53.1 & 73.6 & 50.3 \\
        & Anonymized & 45.8 & 64.2 & 44.1 \\
        & Variation & -7.3 & -9.4 & -6.2 \\
        [2pt]\hdashline\\[-8pt]
        \multirow{3}{*}{KARPA} 
        & Original & 72.3 & 86.4 & 67.2\\
        & Anonymized & 71.8 & 82.3 & 68.7\\
        & Variation & -0.5 & -4.1 & +1.5 \\
        \bottomrule
    \end{tabular}
}
\caption{Performance variation of different methods between original and anonymized WebQSP datasets.}
\vspace{-0.3cm}
\label{an}
\end{table}

\section{Experiments}

In this section, we detail the experimental setup, present our main results, and conduct further analysis to evaluate the performance of KARPA.

\subsection{Experimental Settings}

\paragraph{Datasets and Evaluation Metrics}
We evaluate KARPA on two widely used multi-hop KGQA datasets: WebQuestionSP (WebQSP) \citep{yih2016value} and Complex WebQuestions (CWQ) \citep{talmor2018web}, as well as our newly anonymized version of the WebQSP dataset with placeholders replacing specific details. Evaluation metrics include Accuracy, Hit@1 and F1 score.

\paragraph{Baselines for Comparison}
We compare KARPA against several baselines: (1) LLM-only baselines: IO Prompt \citep{brown2020language} and CoT \citep{wei2022chain} to evaluate LLM reasoning without external knowledge; (2) Training-based methods: KD-CoT \citep{wang2023knowledge}, KG-CoT \citep{zhaokg}, UniKGQA \citep{jiang2022unikgqa}, DECAF \citep{yu2022decaf}, and RoG \citep{luo2023reasoning}, highlighting KARPA's performance without extra training; (3) Direct inference over KGs: ToG \citep{sun2023think} and Interactive-KBQA \citep{xiong2024interactive}, representing the training-free state-of-the-art methods.

\begin{table*}[t]

\begin{center}
\resizebox{0.85\textwidth}{!}{
    \begin{tabular}{lccccccc}
        \toprule
        & & \multicolumn{3}{c}{\textbf{WebQSP}} & \multicolumn{3}{c}{\textbf{CWQ}} \\
        \cmidrule(lr){3-5} \cmidrule(lr){6-8}
        \textbf{Model} & \textbf{Method} & \textbf{Accuracy} & \textbf{Hit@1} & \textbf{F1} & \textbf{Accuracy} & \textbf{Hit@1} & \textbf{F1}  \\
        \midrule
        \multirow{5}{*}{GPT-4o-mini} & CoT  & - & 61.3 & - & - & 49.5 & - \\
        & ToG& 56.4 & 75.2 & 51.6 & 50.2 & 54.0 & 34.5 \\
        & KARPA-B& 67.2 & 82.3 & 61.5 & 66.0 & 72.1 & 57.8 \\
        & KARPA-P& 67.8 & 82.6 & 62.4 & 66.4 & 71.7  & 58.7 \\
        & KARPA-H & \textbf{71.9} & \textbf{85.3} & \textbf{64.5} & \textbf{68.1} & \textbf{73.3} & \textbf{56.5} \\
        [2pt]\hdashline\\[-8pt]
        \multirow{5}{*}{GPT-4o} & CoT  & - & 67.0 & - & - & 52.3 & - \\
        & ToG & 58.6 & 78.5 & 50.9 & 53.3 & 56.8 & 41.9 \\
        & KARPA-B& 73.8 & 85.2 & 67.3 & 65.0 & 70.5 & 55.8 \\
        & KARPA-P& 73.7 & 86.8 & \textbf{69.7} & 69.2 & 74.1 & \textbf{59.8} \\
        & KARPA-H & \textbf{76.1} & \textbf{87.7} & 69.2 & \textbf{69.8} & \textbf{75.3} & 58.4 \\
        [2pt]\hdashline\\[-8pt]
        \multirow{5}{*}{GPT-4} & CoT & - & 66.1 & - & - & 54.7 & - \\
        & ToG* \citep{sun2023think} & - & 82.6 & - & - & 69.5 & -\\
        & KARPA-B& 73.5 & 85.5 & 68.4 & 71.2 & 75.4 & 61.1 \\
        & KARPA-P& 74.1 & 86.8 & 69.3 & 73.4 & 77.9 & \textbf{63.0} \\
        & KARPA-H & \textbf{80.9} & \textbf{91.2} & \textbf{72.1} & \textbf{73.6} & \textbf{78.4} & 61.5 \\
        [2pt]\hdashline\\[-8pt]
        \multirow{5}{*}{Gemini-1.5-Pro} & CoT & - & 65.3 & - & - & 52.1 & -  \\
        & ToG & 62.3 & 78.4 & 52.5 & 51.7 & 57.9 & 40.5 \\
        & KARPA-B& 70.1 & 84.5 & 65.9 & 69.1 & 74.0 & 57.2 \\
        & KARPA-P& 73.8 & 88.0 & 67.4 & 69.6 & 73.5 & \textbf{57.7} \\
        & KARPA-H & \textbf{80.7} & \textbf{90.5} & \textbf{68.6} & \textbf{69.8} & \textbf{75.0} & 54.8 \\
        \bottomrule
\end{tabular}}
\caption{Comparison of KARPA, ToG, and CoT using various LLMs and matching strategies. \textbf{KARPA-B}: Beam search-based matching method with fixed beam width. \textbf{KARPA-P}: Pathfinding-based matching constrained to fixed-length paths. \textbf{KARPA-H}: Heuristic value-based matching allowing similarity calculations across variable-length paths. KARPA consistently outperforms ToG, the prior SOTA for direct KG-based reasoning using LLM.}
\vspace{-0.4cm}
\label{tab:Comparison}
\end{center}
\end{table*}

\paragraph{Experimental Details} 
We test KARPA with various LLMs via API calls. We employ all-MiniLM-L6-v2\citep{reimers-gurevych-2019-sentence} as our embedding model. For each LLM, we randomly select 300 KGs from each datasets to evaluate KARPA's performance, aiming to reduce computational costs. In matching step, we extract 16 top-\(K\) paths with the highest semantic similarity for each candidate paths.

\subsection{Main Results}

\subsubsection{Comparison between Baselines}
We compare KARPA with other approaches in Table~\ref{tab1}.The results show that KARPA significantly outperforms existing baselines across most metrics, achieving state-of-the-art performance. When comparing to the direct answering methods, we demonstrate that leveraging KGs as external knowledge sources enables the LLM to yield superior answers. 

\begin{figure}[t]
\centering
\includegraphics[width=1\columnwidth]{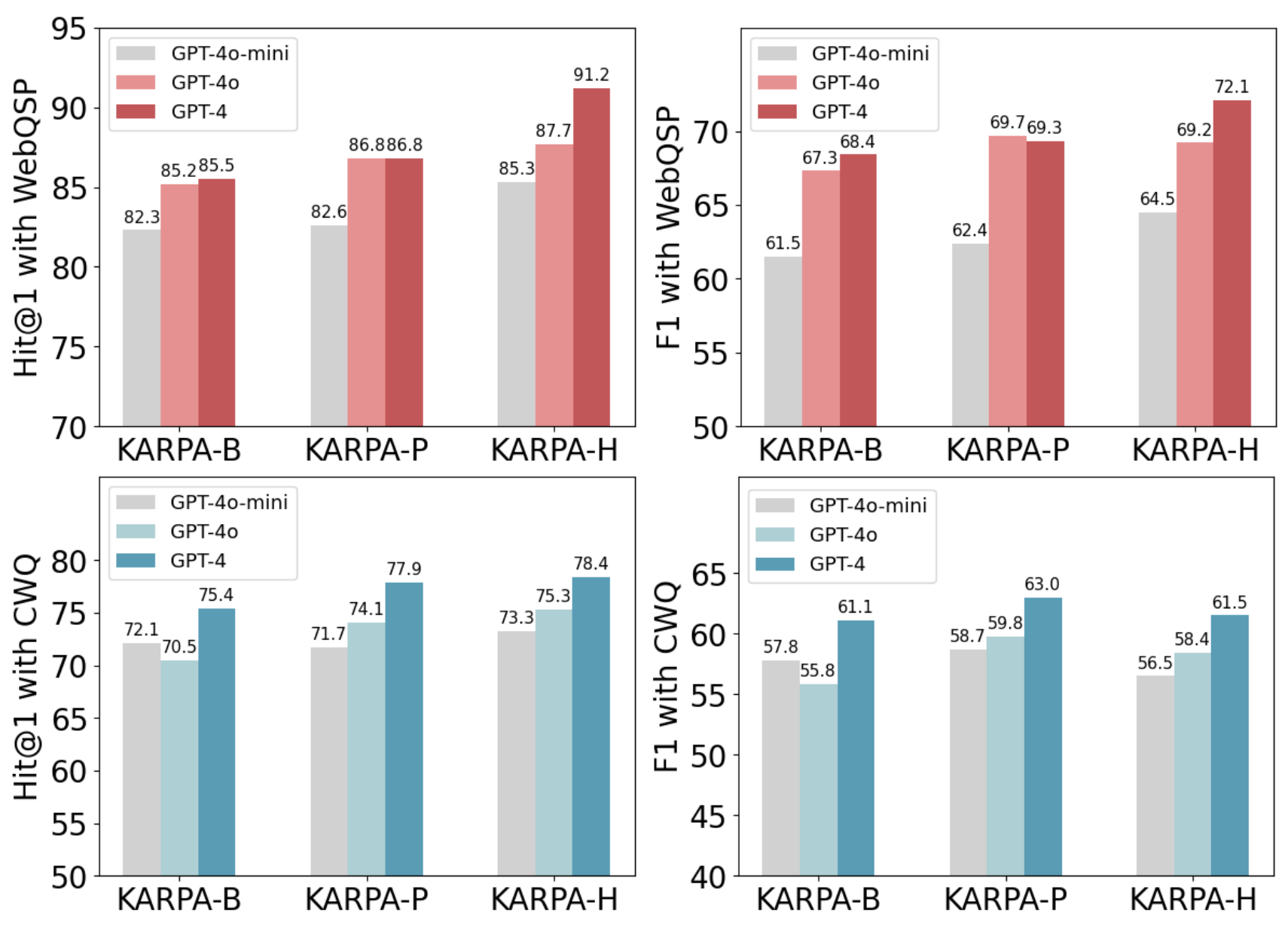}
\caption{Comparison of different matching strategies across various LLMs on Hit@1 and F1 metrics. }
\label{figure3}
\vspace{-0.4cm}
\end{figure} 

In contrast to training-based methods, KARPA is plug-and-play, requiring no additional training while maintaining effective KG-based reasoning. When comparing with inference-based method, which also utilizes LLMs for reasoning over KGs without additional training, KARPA achieves superior results by leveraging LLM's global planning capabilities, enabling the construction of coherent relation paths between topic and answer entities.

To isolate reasoning ability of the LLM from its internal knowledge, we tested KARPA on an anonymized version of the WebQSP dataset, where specific details in questions and answers are replaced with placeholders. For example, question “Where is Jamarcus Russell from? - Mobile.” is transformed into “Where is Person A from? - Location A.” This ensures that the final results are unaffected by LLM's pre-existing knowledge. In Table~\ref{an}, RoG utilize its instruction-tuned LLaMA2-7B for planning and GPT-4o-mini for reasoning, while ToG and KARPA employ GPT-4o-mini for entire pipeline. KARPA shows the smallest performance drop, demonstrating its reliance on reasoning rather than LLM's internal knowledge. RoG exhibits larger decline, highlighting the limitations of instruction-tuned LLMs for unseen KGs. 

\begin{table}[h]
    \centering
    \resizebox{0.80\linewidth}{!}{%
    \begin{tabular}{@{}lcccc@{}}
        \toprule
        \textbf{Dataset} & \textbf{Method} & \textbf{Accuracy} & \textbf{Hit@1} & \textbf{F1} \\
        \midrule
        \multirow{3}{*}{WebQSP} 
        & CoT & - & 41.5 & -\\
        & ToG & 24.6 & 30.2 & 21.9\\
        & KARPA & \textbf{65.6} & \textbf{79.2} & \textbf{58.6}\\
        [2pt]\hdashline\\[-8pt]
        \multirow{3}{*}{CWQ} 
        & CoT & - & 28.3 & -\\
        & ToG & 22.4 & 25.8 & 20.2\\
        & KARPA & \textbf{47.6} & \textbf{52.7} & \textbf{38.8}\\
        \bottomrule
    \end{tabular}
}
\caption{Performance comparison using Qwen2.5-7B.}
\vspace{-0.4cm}
\label{exp_6}
\end{table}

\subsubsection{Performance Across Different LLMs}

We also evaluate ToG and KARPA across various LLMs. As shown in Table~\ref{tab:Comparison}, KARPA consistently outperforms ToG and CoT, regardless of the LLM, by leveraging global planning to construct more logically sound and complete reasoning chains. In contrast, ToG’s reliance on stepwise relation selection limits its effectiveness, as it neglects the LLM's inherent planning capabilities. 

To demonstrate the effectiveness of KARPA on smaller LLMs, we evaluate KARPA with Qwen2.5-7B. Table~\ref{exp_6} shows that KARPA outperforms ToG even with smaller LLMs, demonstrating its robustness and reduced reliance on LLMs' planning abilities. More results are provided in Appendix~\ref{add_exp}.

\section{Analysis and Discussion}

\subsection{Interaction Comparison}

We compare the average number of interactions required by KARPA and ToG across multiple LLMs and datasets. As shown in Table~\ref{tab3}, KARPA reduces interactions by more than half compared to ToG while maintaining higher answer accuracy.

\begin{table}[h]
    \centering
    \resizebox{0.77\linewidth}{!}{%
    \begin{tabular}{@{}lcc@{}}
        \toprule
        Method & WebQSP & CWQ \\
        \midrule
        ToG*\citep{sun2023think} & 11.2 & 14.3 \\
        [2pt]\hdashline\\[-8pt]
        KARPA+GPT-4o-mini & 5.1 & 6.2 \\
        KARPA+GPT-4o & \textbf{4.8} & \textbf{5.3} \\
        KARPA+GPT-4 & 5.5 & 6.0 \\
        \bottomrule
    \end{tabular}
}
\caption{Comparison of LLM call frequency.} 
\vspace{-0.2cm}
\label{tab3}
\end{table}

To further illustrate the reduced reasoning complexity, we compare the average number of input and output tokens for both methods using GPT-4o-mini’s tokenizer. Table~\ref{exp_8} shows KARPA significantly reduces token usage, lowering both reasoning complexity and computational cost.

\begin{table}[htbp]
\centering
\resizebox{1\columnwidth}{!}{
\begin{tabular}{lcc}
    \toprule
    \textbf{WebQSP} & \textbf{Input Tokens/KG} & \textbf{Output Tokens/KG} \\
    \midrule
    ToG    & 6351.5 & 1836.5 \\
    KARPA  & 2465.9 & 1492.3 \\
    \midrule
    \textbf{CWQ} & \textbf{Input Tokens/KG} & \textbf{Output Tokens/KG} \\
    \midrule
    ToG    & 7935.7 & 2931.6 \\
    KARPA  & 3612.1 & 2267.1 \\
    \bottomrule
\end{tabular}}
\caption{Token usage comparison with GPT-4o-mini.}
\vspace{-0.5cm}
\label{exp_8}
\end{table}

\subsection{Ablation Study}

\textbf{Impact of matching methods.} Table~\ref{tab:Comparison} shows that KARPA-H achieves the best matching results, demonstrating the advantage of its flexible and robust performance for KGQA. More results are provided in Appendix~\ref{add_results}.

\textbf{Influence of different LLMs.} Figure~\ref{figure3} shows the impact of LLM capabilities on KARPA’s performance. More powerful LLMs, such as GPT-4, generate better relation paths, leading to more accurate answers \citep{kaplan2020scaling}. With weaker LLMs like GPT-4o-mini, performance declines slightly but still surpasses ToG. This highlights the importance of global planning in KARPA’s design.

\textbf{Influence of Embedding Models.} Table~\ref{exp_4} evaluates KARPA with different embedding models: (1) all-MiniLM-L6-v2 (86MB): Default model of KARPA. (2) all-mpnet-base-v2 (417MB): More powerful embedding model. (3) paraphrase-multilingual-MiniLM-L12-v2 (448MB): Supports embedding between multiple languages. The results demonstrate that KARPA’s performance remains stable across different embedding models.

\begin{table}[htbp]
\centering
\resizebox{1\columnwidth}{!}{
\begin{tabular}{lccc}
    \toprule
    \textbf{Embedding Model} & \textbf{Accuracy} & \textbf{Hit@1} & \textbf{F1} \\
    \midrule
    all-MiniLM-L6-v2 & 72.3 & \textbf{86.4} & 67.2 \\
    all-mpnet-base-v2 & \textbf{74.5} & 86.1 & \textbf{68.6} \\
    multilingual-MiniLM-L12-v2 & 74.1 & 85.3 & 68.3 \\
    \bottomrule
\end{tabular}}
\caption{KARPA with different embedding models.}
\label{exp_4}
\end{table}

\subsection{Discussion}

KARPA requires fewer interactions and token usage comparing to ToG, while still outperforming ToG even when using smaller LLMs. This efficiency stems from KARPA’s ability to generate complete reasoning chains, reducing the need for stepwise interactions in other methods. Methods like ToG impose heavy computational burdens by evaluating hundreds or even thousands of adjacent relations at each step, whereas KARPA’s global planning aligns better with human-like reasoning.

Table~\ref{exp_4} demonstrates that KARPA performs robustly across embedding models. Its pre-planned paths are distinctive and semantically aligned with correct reasoning paths, making even lightweight embedding models sufficient for path matching.

\section{Conclusion}

In this paper, we propose KARPA, a novel framework designed to enhance LLM-based KGQA by utilizing the global planning and reasoning capabilities of LLMs. KARPA addresses key limitations of existing methods, achieving superior accuracy and efficiency through its pre-planning, matching and reasoning processes. Our experiments show that KARPA consistently outperforms state-of-the-art methods across multiple datasets. Its training-free design allows seamless integration with various LLMs, making it broadly applicable to different KGQA tasks. By optimizing LLM-KG interactions, KARPA enhances reasoning efficiency and effectiveness, highlighting its potential as a robust approach for future RAG systems.

\section{Limitations}

Although KARPA effectively reduces the reliance on the capacity of LLMs, its performance is still influenced by the reasoning and planning capabilities of the LLMs themselves. In situations where weaker LLMs are used, KARPA’s performance may degrade due to LLMs' limited ability to generate logically coherent paths or perform intricate reasoning tasks. In our future work, we aim to enhance KARPA's performance on weaker LLMs, ensuring that KARPA remains effective across a broader range of LLMs with varying levels of reasoning and planning capabilities.

\bibliography{custom}

\appendix

\section{Algorithm for KARPA}

In this section, we present the pseudo-code for the Knowledge graph Assisted Reasoning Path Aggregation (KARPA) framework, as shown in Algorithm~\ref{alg:karpa}. The pseudo-code outlines the key components of our approach, including the pre-planning, matching, and reasoning phases. It demonstrates the interaction between the large language model (LLM) and the embedding model in generating, matching, and refining relation paths, which are crucial for improving LLM-based KGQA tasks.

\begin{algorithm}
    \caption{KARPA Framework}
    \label{alg:karpa}
    \KwIn{Question $Q$, Topic entity $e_t$, Knowledge Graph $KG$}
    \KwOut{Answers $E_a$}
    \textbf{Pre-Planning Phase:}\\
    Generate initial paths $P_i = \{p_1, p_2, \dots, p_m\}$ using $LLM(Q, e_t)$\;
    \For{each path $p_i = (r_1^i, r_2^i, \dots, r_{n_i}^i)$}
    {
        Decompose $p_i$ into relation list $R_i = \{r_1^i, r_2^i, \dots, r_{n_i}^i\}$\;
        \For{each relation $r_j^i$ in $R_i$}{
            Retrieve top-$K$ similar relations $R_j^i = \text{Top-K}(\text{sim}(\mathbf{r_j^i}, \mathbf{r}))$\;
        }
    }
    Re-plan relation paths $P_{cand} = \text{LLM}(Q, R_j^i)$ based on extracted relations $R_j^i$\;
    \textbf{Matching Phase:}\\
    Extract relation paths $P_r$ with length $L \in len(P_{cand})$\;
    \For{each path $p$ in $P_{cand}$}{
        Compute similarity between paths using heuristic value $P_{matched} = Heuristic (\text{sim}(p, p_r), p_r \in P_r)$\;
        Extract top-$K$ similar paths $P = \text{Top-K}(P_{matched})$\;
    }
    \textbf{Reasoning Phase:}\\
    Combine relation paths $P_{matched} = \{r_1, r_2, \dots, r_n\}$ with $e_t$, $e_a$ into prompt\;
    Predict final answer $E_a = \text{LLM}(Q, P_{matched}, e_t, e_a)$\;
    \Return $E_a$
\end{algorithm}

\begin{table*}[t]
\begin{center}
\resizebox{0.75\textwidth}{!}{
    \begin{tabular}{lccccccc}
        \toprule
        & & \multicolumn{4}{c}{\textbf{WebQSP}} \\
        \cmidrule(lr){3-6} 
        Model Tpye & Method & Accuracy & Hit@1 & F1 & Precision \\
        \midrule
        \multirow{3}{*}{GPT-4o-mini} & KARPA-B  & 67.2 & 82.3 & 61.5 & 64.1 \\
        & KARPA-P & 67.8 & 82.6 & 62.4 & 64.9 \\
        & KARPA-H & \textbf{71.9} & \textbf{85.3} & \textbf{64.5} & \textbf{65.9} \\
        \midrule
        \multirow{3}{*}{GPT-4o} & KARPA-B  & 73.8 & 85.2 & 67.3 & 72.3 \\
        & KARPA-P& 73.7 & 86.8 & \textbf{69.7} & 70.5 \\
        & KARPA-H & \textbf{76.1} & \textbf{87.7} & 69.2 & \textbf{71.5} \\
        \midrule
        \multirow{3}{*}{GPT-4} & KARPA-B & 73.5 & 85.5 & 68.4 & 71.7 \\
        & KARPA-P & 74.1 & 86.8 & 69.3 & \textbf{73.6} \\
        & KARPA-H & \textbf{80.9} & \textbf{91.2} & \textbf{72.1} & 73.1 \\
        \midrule
        \multirow{3}{*}{DeepSeek-V2.5} & KARPA-B & 71.8 & 84.0 & 63.1 & 65.9 \\
        & KARPA-P & 73.4 & 85.3 & 64.1 & 66.3\\
        & KARPA-H & \textbf{78.1} & \textbf{88.4} & \textbf{68.7} & \textbf{67.6}\\
        \midrule
        \multirow{3}{*}{Gemini-1.5-Pro} & KARPA-B & 70.1 & 84.5 & 65.9 & 64.7 \\
        & KARPA-P & 73.8 & 88.0 & 67.4 & 66.1\\
        & KARPA-H & \textbf{80.7} & \textbf{90.5} & \textbf{68.6} & \textbf{67.8}\\
        \midrule
        \multirow{3}{*}{Claude-3.5-Sonnet} & KARPA-B & 75.1 & 85.7 & 66.0 & 67.6 \\
        & KARPA-P & 80.4 & 89.0 & 69.7 & \textbf{70.4} \\
        & KARPA-H & \textbf{82.6} & \textbf{89.5} & \textbf{69.7} & 69.1\\
        \bottomrule
\end{tabular}}
\caption{Performance of KARPA with different matching strategies (KARPA-B, KARPA-P, and KARPA-H) and LLMs on the WebQSP dataset.}
\label{tab5}
\end{center}
\end{table*}

\begin{table*}[t]
\begin{center}
\resizebox{0.75\textwidth}{!}{
    \begin{tabular}{lccccccc}
        \toprule
        & & \multicolumn{4}{c}{\textbf{CWQ}} \\
        \cmidrule(lr){3-6} 
        Model Tpye & Method & Accuracy & Hit@1 & F1 & Precision \\
        \midrule
        \multirow{3}{*}{GPT-4o-mini} & KARPA-B  & 66.0 & 72.1 & 57.8 & 58.6 \\
        & KARPA-P & 66.4 & 71.7 & \textbf{58.7} & \textbf{59.8} \\
        & KARPA-H & \textbf{68.1} & \textbf{73.3} & 56.5 & 55.1 \\
        \midrule
        \multirow{3}{*}{GPT-4o} & KARPA-B  & 65.0 & 70.5 & 55.8 & 57.8 \\
        & KARPA-P& 69.2 & 74.1 & \textbf{59.8} & 58.4 \\
        & KARPA-H & \textbf{69.8} & \textbf{75.3} & 58.4 & \textbf{59.5} \\
        \midrule
        \multirow{3}{*}{GPT-4} & KARPA-B & 71.2 & 75.4 & 61.1 & 62.7 \\
        & KARPA-P & 73.4 & 77.9 & \textbf{63.0} & 62.5 \\
        & KARPA-H & \textbf{73.6} & \textbf{78.4} & 61.5 & \textbf{63.1} \\
        \midrule
        \multirow{3}{*}{DeepSeek-V2.5} & KARPA-B & 61.6 & 63.2 & 48.4 & 50.1 \\
        & KARPA-P & 60.9 & 63.0 & 51.8 & 52.6\\
        & KARPA-H & \textbf{62.6} & \textbf{64.1} & \textbf{51.9} & \textbf{53.5}\\
        \midrule
        \multirow{3}{*}{Gemini-1.5-Pro} & KARPA-B & 69.1 & 74.0 & 57.2 & 59.5 \\
        & KARPA-P & 69.6 & 73.5 & \textbf{57.7} & \textbf{60.3}\\
        & KARPA-H & \textbf{69.8} & \textbf{75.0} & 54.8 & 55.8\\
        \midrule
        \multirow{3}{*}{Claude-3.5-Sonnet} & KARPA-B & 62.8 & 65.7 & 49.6 & 52.1 \\
        & KARPA-P & 61.5 & 64.3 & 52.9 & 55.5 \\
        & KARPA-H & \textbf{70.6} & \textbf{73.7} & \textbf{54.9} & \textbf{56.9}\\
        \bottomrule
\end{tabular}}
\caption{Performance of KARPA with different matching strategies (KARPA-B, KARPA-P, and KARPA-H) and LLMs on the CWQ dataset.}
\label{tab6}
\end{center}
\end{table*}

\section{Implementation Details}
 
\paragraph{Model Invocation.} 
KARPA is tested with LLMs such as GPT-4~\citep{openai2023gpt4}, GPT-4o~\citep{openai2024gpt4o}, GPT-4o-mini, Claude-3.5-Sonnet~\citep{claude35addendum}, Gemini-1.5-pro~\citep{geminiteam2024gemini15unlockingmultimodal}, and other LLMs through API calls. These LLMs are queried dynamically throughout the experimental pipeline to perform pre-planning, matching, and reasoning steps.

\paragraph{Experimental Setup.} 
During the pre-planning stage, the initial paths generated by the LLM are decomposed and stored, along with the query, into a list. For each element in this list, we extract the top-k relations, where the total number of extracted relations does not exceed 30. These relations are semantically closest to the elements based on the LLM's initial output. 

In the matching step, KARPA selects the top 16 relation paths with the highest similarity for each initial relation path. These paths serve as candidate paths for reasoning step. In the reasoning step, we limit the number of candidate paths input to the LLM at one time to a maximum of 8, ensuring that the reasoning process remains manageable and focused on the most relevant paths.

\paragraph{Answer Evaluation.} 
To determine if the LLM correctly answers the question, KARPA enforces a specific output format. The final answer must be enclosed in curly brackets in the LLM’s output. We consider an answer correct only when the tail entities of the reasoning paths match the text enclosed within the curly brackets in the LLM's output. For CoT, we consider an answer correct if the LLM's response contains the correct answer entities. This difference reflects the distinct reasoning and output expectations between KARPA and CoT.

\section{Additional Results}\label{add_results}

In this section, we present additional experimental results to further evaluate the performance of KARPA when using different matching methods: KARPA-B (beam search-based matching strategy), KARPA-P (pathfinding-based matching strategy), and KARPA-H (heuristic value-based matching strategy). We conduct these experiments across various LLMs, analyzing the effectiveness of each matching strategy in conjunction with different LLMs. These results provide a deeper insight into how different matching mechanisms impact the overall performance of KARPA, showcasing the versatility and adaptability of our approach under varying model conditions.

The results presented in Table~\ref{tab5} and Table~\ref{tab6} consistently demonstrate the superior performance of KARPA-H (heuristic value-based matching) compared to the other two matching strategies, KARPA-B (beam search-based) and KARPA-P (pathfinding-based), across different LLMs and datasets (WebQSP and CWQ).

In the majority of LLMs, KARPA-H outperforms the other methods in most metrics. This suggests that KARPA-H is more effective at extracting the correct relation paths, which in turn leads to more accurate and contextually relevant answers. These results highlight KARPA-H as the most robust and reliable matching method among the three, reinforcing its advantage in handling complex KG-based reasoning tasks. 

\section{Additional Experiments}\label{add_exp}

In this section, we provide additional experiments to validate KARPA's performance from different perspectives.

To demonstrate that KARPA has better generalization capabilities than methods based on instruction-tuned LLMs, we conducted an experiment using GPT-4o-mini with a modified version of the WebQSP dataset. Specifically, we slightly alter the questions in WebQSP dataset while preserving their original meaning, using the prompt: "Please revise the question to make it more clear, but the original meaning of the question and the corresponding answers remain unchanged." We test RoG using its instruction-tuned LLaMa2-Chat-7B from in the planning step and GPT-4o-mini for reasoning. In KARPA, we use GPT-4o-mini for both pre-planning and reasoning steps. 

\begin{table*}[h]
\centering
\begin{tabular}{c|cccccccc}
\hline
\textbf{Question} & \textbf{Method} & \textbf{Accuracy} & \textbf{Hit@1} & \textbf{F1} & \textbf{Method} & \textbf{Accuracy} & \textbf{Hit@1} & \textbf{F1} \\ \hline
Origin   & RoG  & 67.6  & 84.1  & 69.7 & KARPA  & 73.1  & 85.4  & 68.1 \\
Revised  & RoG  & 63.5  & 74.3  & 64.1 & KARPA  & 72.6  & 84.5  & 68.9 \\
Variation& RoG  & -4.1  & -9.8  & -5.6 & KARPA  & -0.5  & -0.9  & +0.8 \\ \hline
\end{tabular}
\caption{Comparison of RoG and KARPA on the WebQSP dataset with original and revised questions.}
\label{exp_1}
\end{table*}

The results in Table~\ref{exp_1} show that KARPA’s performance remains consistent and robust to question modifications, while RoG’s performance drops due to path mismatches. This further highlights the advantage of KARPA's training-free framework, maintaining superior robustness and adaptability across all KGs.

To demonstrate the effectiveness of KARPA with smaller LLMs, we conduct experiments with Qwen2.5-7B and Qwen2.5-14B as the LLM backbones for KARPA. The results in Table~\ref{exp_2} demonstrate that KARPA consistently outperforms stepwise direct inference baselines such as ToG, even when using smaller LLMs. This reinforces the robustness and adaptability of our method across different LLM scales.

\begin{table}[h]
    \centering
    \resizebox{0.95\linewidth}{!}{%
    \begin{tabular}{@{}lcccc@{}}
        \toprule
         &  & \multicolumn{3}{c}{\textbf{WebQSP}} \\
        \cmidrule(l){3-5}
        \textbf{Model Type} & \textbf{Method} & \textbf{Accuracy} & \textbf{Hit@1} & \textbf{F1} \\
        \midrule
        \multirow{3}{*}{Qwen2.5-7B} 
        & CoT & - & 41.5 & -\\
        & ToG & 24.6 & 30.2 & 21.9\\
        & KARPA & \textbf{65.6} & \textbf{79.2} & \textbf{58.6}\\
        [2pt]\hdashline\\[-8pt]
        \multirow{3}{*}{Qwen2.5-14B} 
        & CoT & - & 49.6 & - \\
        & ToG & 45.0 & 55.9 & 42.7\\
        & KARPA & \textbf{72.6} & \textbf{84.1} & \textbf{65.0}\\
        \midrule
        &  &  & \textbf{CWQ} & \\
        \midrule
        \multirow{3}{*}{Qwen2.5-7B} 
        & CoT & - & 28.3 & -\\
        & ToG & 22.4 & 25.8 & 20.2\\
        & KARPA & \textbf{47.6} & \textbf{52.7} & \textbf{38.8}\\
        [2pt]\hdashline\\[-8pt]
        \multirow{3}{*}{Qwen2.5-14B} 
        & CoT & - & 31.2 & - \\
        & ToG & 30.2 & 36.6 & 29.5\\
        & KARPA & \textbf{51.5} & \textbf{57.9} & \textbf{41.6}\\
        \bottomrule
    \end{tabular}
}
\caption{Performance comparison of different methods on WebQSP and CWQ datasets using smaller LLMs.}
\label{exp_2}
\end{table}

Also, the results in Table~\ref{exp_2} show that KARPA can perform well with LLMs that have weaker planning and reasoning capabilities, further highlighting KARPA’s robustness and its reduced dependence on the LLM’s planning and reasoning abilities compared to other inference-based methods.

To quantify the impact of the re-planning step, we provide an ablation study that removes the re-planning step from the pre-planning stage. The re-planning step is designed to handle mismatches between LLMs and KGs. In re-planning step, the extracted relations are used to refine and re-plan candidate paths. This guarantees that the candidate paths are both logically coherent and aligned with the KG.

\begin{table*}[htbp]
\centering
\begin{tabular}{lcccccc}
    \toprule
    & \multicolumn{3}{c}{\textbf{WebQSP}} & \multicolumn{3}{c}{\textbf{CWQ}} \\
    \cmidrule(lr){2-4} \cmidrule(lr){5-7}
    \textbf{Pre-Planning} & \textbf{Accuracy} & \textbf{Hit@1} & \textbf{F1} & \textbf{Accuracy} & \textbf{Hit@1} & \textbf{F1} \\
    \midrule
    Origin & 72.3 & 86.4 & 67.2 & 64.6 & 67.7 & 55.1 \\
    Remove Re-Planning Step & 64.1 & 79.6 & 61.5 & 54.3 & 59.5 & 47.1 \\
    \bottomrule
\end{tabular}
\caption{Ablation study of removing re-planning step from the pre-planning stage.}
\label{exp_7}
\end{table*}

The results in Table~\ref{exp_7} show that the re-planning step is crucial for KARPA's performance. Additionally, in the matching step, KARPA employs semantic similarity as the cost function for pathfinding algorithms. This ensures that the final reasoning paths selected not only exist in the KG but are also semantically closest to the paths generated by the LLM, thereby maintaining the validity of the LLM's output across diverse query problems.

\begin{table*}[htbp]
\centering
\begin{tabular}{lcccccc}
    \toprule
    & \multicolumn{3}{c}{\textbf{WebQSP}} & \multicolumn{3}{c}{\textbf{CWQ}} \\
    \cmidrule(lr){2-4} \cmidrule(lr){5-7}
    \textbf{Language} & \textbf{Accuracy} & \textbf{Hit@1} & \textbf{F1} & \textbf{Accuracy} & \textbf{Hit@1} & \textbf{F1} \\
    \midrule
    English-English  & 74.1 & 85.3 & 68.3 & 65.3 & 69.5 & 55.4 \\
    Chinese-English  & 74.6 & 84.5 & 67.6 & 63.1 & 68.0 & 54.2 \\
    \bottomrule
\end{tabular}
\caption{Performance comparison of different languages using a multilingual embedding model.}
\label{exp_9}
\end{table*}

In multilingual scenarios, KARPA can effectively address this problem by using multilingual embedding models. For instance, in a multilingual setting, we test KARPA with paraphrase-multilingual-MiniLM-L12-v2, a multilingual embedding model. In the multilingual experiment, we use GPT-4o-mini to generate relation paths in Chinese, and then use the multilingual embedding model to calculate the semantic similarity between the candidate paths and paths in the KG.

These results in Table~\ref{exp_9} demonstrate that with a multilingual embedding model, KARPA performs effectively across languages, maintaining its robustness. They also indicate that language variations do not significantly impact KARPA’s performance.

To demonstrate the necessity of extending relation paths with different lengths, we restrict the matching step to use only single-relation candidate paths provided by the LLM during re-planning step, and compare the performance of the heuristic value-based matching method (KARPA-H) with the pathfinding-based matching method (KARPA-P) using GPT-4o-mini.

\begin{table*}[htbp]
\centering
\begin{tabular}{llcccccc}
    \toprule
    & & \multicolumn{3}{c}{\textbf{WebQSP}} & \multicolumn{3}{c}{\textbf{CWQ}} \\
    \cmidrule(lr){3-5} \cmidrule(lr){6-8}
    \textbf{Candidate Path} & \textbf{Method} & \textbf{Accuracy} & \textbf{Hit@1} & \textbf{F1} & \textbf{Accuracy} & \textbf{Hit@1} & \textbf{F1}  \\
    \hline
    Original Paths         & KARPA-P & 66.0 & 81.2 & 63.8 & 61.0 & 64.5 & 53.4 \\
    Original Paths         & KARPA-H & 72.3 & 86.4 & 67.2 & 64.6 & 67.7 & 55.1 \\
    Single-Relation Paths  & KARPA-P & 63.6 & 77.3 & 60.7 & 40.5 & 43.9 & 39.3 \\
    Single-Relation Paths  & KARPA-H & 71.4 & 85.5 & 68.9 & 55.1 & 59.6 & 47.4 \\
    \hline
\end{tabular}
\caption{Performance of KARPA-P and KARPA-H using different candidate paths on the WebQSP and CWQ datasets.}
\label{exp_10}
\end{table*}

The results in the Table~\ref{exp_10} demonstrate that the heuristic value-based matching method outperforms pathfinding-based matching methods in such scenarios, as it effectively addresses the semantic similarity issues that arise from differing path lengths. Moreover, as the questions in the CWQ dataset generally require longer reasoning paths compared to WebQSP, both methods exhibit a more significant decline in various metrics on CWQ. However, the heuristic value-based method shows a less pronounced drop compared to pathfinding-based methods, further demonstrating its superiority.

To validate the performance of KARPA on KGs outside the training scope, we compare KARPA with Chain-of-Thought (CoT) reasoning, where the LLM directly relies on its internal knowledge to answer questions. Using open source LLMs such as Qwen2.5-7B, Qwen2.5-14B and Qwen2.5-72B (with limited stored knowledge), we observe that CoT performance drops significantly on KGQA tasks while KARPA maintains strong performance.

\begin{table*}[htbp]
\centering
\begin{tabular}{llcccccc}
    \toprule
    & & \multicolumn{3}{c}{\textbf{WebQSP}} & \multicolumn{3}{c}{\textbf{CWQ}} \\
    \cmidrule(lr){3-5} \cmidrule(lr){6-8}
    \textbf{Base-Model} & \textbf{Method} & \textbf{Accuracy} & \textbf{Hit@1} & \textbf{F1} & \textbf{Accuracy} & \textbf{Hit@1} & \textbf{F1}  \\
    \midrule
    \multirow{3}{*}{Qwen2.5-7B} & CoT    & -     & 41.5 & -     & -     & 28.3 & -     \\
                               & KARPA  & 65.6  & 79.2 & 58.6  & 47.6  & 52.7 & 38.8  \\
                               & \textbf{Gain} & -     & \textbf{+37.7} & -     & -     & +24.4 & -     \\
    \hdashline \\[-8pt]
    \multirow{3}{*}{Qwen2.5-14B} & CoT    & -     & 49.6 & -     & -     & 31.2 & -     \\
                               & KARPA  & 72.6  & 84.1 & 65.0  & 51.5  & 57.9 & 41.6  \\
                               & \textbf{Gain} & -     & +34.5 & -     & -     & \textbf{+26.7} & -     \\
    \hdashline \\[-8pt]
    \multirow{3}{*}{Qwen2.5-72B} & CoT    & -     & 56.9 & -     & -     & 40.5 & -     \\
                               & KARPA  & 73.2  & 86.0 & 64.5  & 61.1  & 63.6 & 52.7  \\
                               & \textbf{Gain} & -     & +29.1 & -     & -     & +23.1 & -     \\
    \hline
\end{tabular}
\caption{Performance comparison of CoT and KARPA methods across different base models (Qwen2.5-7B, Qwen2.5-14B, Qwen2.5-72B) on WebQSP and CWQ datasets.}
\label{exp_11}
\end{table*}

The results in Table~\ref{exp_11} highlight KARPA's ability to operate effectively on unseen KGs by focusing on reasoning and planning rather than leveraging the LLM’s pre-existing knowledge. The results also show that KARPA maintained strong performance, even as the LLM’s stored knowledge was significantly reduced. This means that even if the LLM does not have ample prior knowledge about a specific domain, KARPA can still leverage the LLM's reasoning and planning capabilities to construct reasoning chains to find the correct answers within the KG.

To demonstrate the effectiveness of KARPA in noisy KGs and specialized domains, we conduct an experiment introducing noise into the KG. For WebQSP and CWQ samples with reasoning paths longer than one, we randomly shuffle the neighboring relations of topic entity and then compared the performance of KARPA and ToG using GPT-4o-mini.

The results in Table~\ref{exp_12} show that KARPA experiences a slight drop in performance, demonstrating its resilience to noisy relations. ToG shows a more significant decline, highlighting the limitations of traditional KGQA methods in noisy environments.

\section{Further Discussion}

\subsection{Effectiveness Beyond KGQA Tasks}

While KARPA is currently designed to address challenges in KGQA tasks, following the settings of prior works such as RoG and ToG, its methodology is generalizable to other knowledge-intensive tasks.

KARPA’s core idea lies in letting LLMs generate complete reasoning chains instead of disrupting reasoning continuity with step-by-step searching. This approach mimics human reasoning processes and enhances reasoning efficiency. For example, in knowledge-intensive task such as the retrieval of academic papers, KARPA could generate reasoning chains like “research field → target journal/conference → specific keywords”, and then retrieve the corresponding paper using semantic similarity. When extracting information from books, the reasoning chain like “book title → relevant chapter → relevant paragraphs” could streamline the information extraction. This reasoning-chain generation aligns with human thought processes, making it both intuitive and adaptable to diverse knowledge-intensive tasks.

\subsection{Incorporating User Feedback Mechanisms}

KARPA’s architecture is inherently well-suited to incorporating user feedback mechanisms due to its design of generating complete reasoning paths. We provide a potential extension here:

\begin{itemize}
    \item Initial Path Generation: KARPA generates an initial reasoning path based on the user query.
    \item Ambiguity Threshold: Using our semantic similarity-based matching method, we match the LLM-generated path with paths within the KG. If the similarity score reaches a certain ambiguity threshold, the query is considered clear; if the similarity score falls below that threshold, we identify the query as potentially ambiguous.
    \item User Feedback: If the similarity score reaches the threshold, we can provide the user with the retrieved answers. If the score falls below the threshold, we could present the extracted reasoning paths to the user for review and request further clarification or refinement of the query.
    \item Refinement and Rematching: Based on user feedback, KARPA could adjust the reasoning path and re-run the matching process to generate more accurate results.
\end{itemize}

Through the steps outlined above, KARPA can establish a comprehensive user feedback mechanism, which enhances the precision of queries based on ongoing user feedback.

\begin{table*}[htbp]
\centering
\begin{tabular}{llcccccc}
    \toprule
    & & \multicolumn{3}{c}{\textbf{WebQSP}} & \multicolumn{3}{c}{\textbf{CWQ}} \\
    \cmidrule(lr){3-5} \cmidrule(lr){6-8}
    \textbf{Knowledge Graphs} & \textbf{Method} & \textbf{Accuracy} & \textbf{Hit@1} & \textbf{F1} & \textbf{Accuracy} & \textbf{Hit@1} & \textbf{F1}  \\
    \midrule
    Original KGs     & ToG   & 54.2  & 72.8  & 50.3  & 47.6  & 52.5  & 39.1  \\
    Shuffled KGs     & ToG   & 32.7  & 48.2  & 30.1  & 23.3  & 26.7  & 20.9  \\
    \textbf{Variation}        & ToG   & -21.5 & -24.6 & -20.2 & -24.3 & -25.8 & -18.2 \\
    \hdashline \\[-8pt]
    Original KGs     & KARPA & 72.3  & 86.4  & 67.2  & 64.6  & 67.7  & 55.1  \\
    Shuffled KGs     & KARPA & 70.7  & 84.1  & 64.5  & 56.0  & 61.3  & 51.5  \\
    \textbf{Variation}        & KARPA & -1.6  & -2.3  & -2.7  & -8.6  & -6.4  & -3.6  \\
    \hline
\end{tabular}
\caption{Comparison of performance between original and shuffled KGs for ToG and KARPA methods on WebQSP and CWQ datasets.}
\label{exp_12}
\end{table*}

\section{Detailed Related Work}

\subsection{Prompt-Based Question Answering Using Internal Knowledge}
In the field of large language models (LLMs), researchers explore how to combine internal knowledge with external information to enhance reasoning abilities. Existing models utilize a vast internal knowledge base and achieve significant progress in reasoning tasks. To further optimize these capabilities, researchers propose various prompt-based methods, such as Chain of Thought (CoT) \citep{li2023chain} prompting. This method breaks down complex tasks into manageable steps, promoting structured reasoning and excelling in mathematical and logical reasoning. Building on CoT, researchers also develop variants like Auto-CoT \citep{zhang2022automatic}, Zero-Shot-CoT \citep{kojima2022large}, Complex-CoT \citep{fu2022complexity}, and new frameworks such as Tree of Thoughts (ToT) \citep{yao2024tree}, which further expand the application range of LLMs.

Additionally, with regard to the ``decoding" problem of the reasoning process, Self-consistency CoT \citep{wang2022self} serves as a representative method. It generates multiple reasoning paths through manually designed prompts and employs a ``majority voting" mechanism to identify the ``most consistent" path, thereby enhancing CoT performance. CoT verification 
\citep{weng2022large} is another important research direction that allows models to self-verify the correctness of their answers through multiple rounds of reasoning. Self-Verification samples multiple candidate reasoning paths and ranks them based on whether the conditions satisfy the conclusions. Recently, OpenAI launches the o1 series models, marking a significant advancement in LLM reasoning abilities, allowing models to develop extensive internal chains of thought and further tap into their reasoning potential.

\subsection{Embedding models and Embedding-based methods.}

Embedding models have revolutionized how we represent and understand text by converting words and sentences into dense vector representations \citep{mikolov2013distributed}. These embedding models capture the semantic meaning of the text, enabling models to effectively measure the similarity and relationships between different texts. In recent years, significant progress has been made in the field of text embeddings, largely due to the emergence of pre-trained language models \citep{vaswani2017attention}. Models like BERT and its variants have become fundamental tools for efficiently encoding the underlying semantics of data. Key advancements in contrastive learning, particularly improvements in negative sampling and knowledge distillation applications, also contribute significantly to the progress in this field. As a result, there is a growing trend to develop universal embedding models that can uniformly support a variety of applications, ranging from information retrieval to natural language processing tasks.

\subsection{Knowledge Graphs and Retrieval-Augmented Methods.}

Knowledge graphs and retrieval-augmented generation (RAG) \citep{lewis2020retrieval} play a crucial role in enhancing various downstream tasks, such as question answering, text generation, and information retrieval. Early research \cite{sun2018open} uses random walk algorithms to retrieve information from knowledge graphs. Subsequent studies \cite{li2023graph,yu2021kg} employ BM25 and DPR algorithms for knowledge graph-based information retrieval, further improving the performance of LLMs. UniKGQA \cite{jiang2022unikgqa} integrates the retrieval process with LLMs to achieve state-of-the-art performance in knowledge graph question-answering tasks. KELP utilizes an embedding model to filter reasoning paths from the KG. However, it does not leverage the reasoning capabilities of LLMs and is limited to reasoning paths within a 2-hop range, restricting its applicability to more complex queries. KnowledgeNavigator \citep{guo2024knowledgenavigator} employs an iterative process where the LLM retrieves and filters relevant knowledge directly from the KG. GraphRAG \cite{edge2024local} designs a powerful process that extracts structured data from unstructured text using LLMs. These studies collectively demonstrate that information retrieved from knowledge graphs significantly enhances the reasoning capabilities of LLMs. 

\section{Datasets}
We adopt two widely-used multi-hop KGQA datasets in our work. Table~\ref{dataset} below gives detailed statistical information for both datasets.
\begin{itemize}
  \item \textbf{WebQuestionsSP (WebQSP)}~\citep{yih2016value}  is a knowledge base Q\&A dataset containing 4737 questions requiring up to 2-hop reasoning on the KG Freebase~\citep{bollacker2008freebase}, designed to improve the performance of Q\&A systems through semantic parsing.
  \item \textbf{Complex WebQuestion (CWQ)}~\citep{talmor2018web} is extended based on the WebQSP dataset that require up to 4-hop reasoning on the KG Freebase~\citep{bollacker2008freebase} to solve more complex Q\&A tasks.
\end{itemize}

\begin{table}[h]
    \centering
    \begin{tabular}{lcc}
    \toprule
    \textbf{Statistics} & \textbf{WebQSP} & \textbf{CWQ} \\
    \midrule
    \rowcolor{blue!10} 
    \multicolumn{3}{c}{\textbf{Dataset Split}} \\
    \midrule
    Train & 2,826  & 27,639 \\
    Test  & 1,628  & 3,531  \\

    \midrule
    \rowcolor{blue!10}
    \multicolumn{3}{c}{\textbf{Question Hop Distribution}} \\
    \midrule
    1 hop  & 65.49\% & 40.91\% \\
    2 hop  & 34.51\% & 38.34\% \\
    $\geq$ 3 hop  & 0.00\% & 20.75\% \\
    
    \midrule
    \rowcolor{blue!10}
    \multicolumn{3}{c}{\textbf{Answer Counts Distribution}} \\
    \midrule
    Ans = 1 & 51.2\%  & 70.6\%  \\
    2 $\leq$ Ans $\leq$ 4 & 27.4\%  & 19.4\%  \\
    5 $\leq$ Ans $\leq$ 9 & 8.3\%   & 6.0\%   \\
    Ans $\geq$ 10    & 12.1\%  & 4.0\%   \\
    \bottomrule
    \end{tabular}
    \caption{Comprehensive Statistics of Datasets.}
    \label{dataset}
\end{table}

\section{Baselines}

We consider the following baseline methods for performance comparison:
\begin{itemize}
\item \textbf{IO Prompt}: Directly query large language models (LLMs) for answers without relying on external sources of information or additional reasoning processes.
\item \textbf{CoT Prompt}: Utilizing Chain-of-Thought prompting with LLMs to facilitate reasoning involves guiding the LLM through a step-by-step process, where each step reflects the logical sequence of human reasoning. 
\item \textbf{LLM-Based KGQA Methods}:

    \textbf{KD-CoT}~\citep{wang2023knowledge} interacts with external knowledge to verify and amend the reasoning paths within the Chain-of-Thought (CoT), effectively overcoming issues of hallucinations and error propagation. It structures the CoT reasoning process of LLMs into a formatted multi-round QA approach. In each round, LLMs interact with a QA system that retrieves external knowledge, constructing more reliable reasoning paths based on the precise answers retrieved, thereby enhancing the accuracy and credibility of reasoning.
    
    \textbf{UniKGQA}~\citep{jiang2022unikgqa} unifies retrieval and reasoning in both model architecture and parameter learning by designing a shared pre-training task based on question-relation matching and applying fine-tuning strategies to optimize the retrieval and reasoning processes. It includes two main modules: a semantic matching module based on a pre-trained language model (PLM) for question-relation semantic matching, and a matching information propagation module that spreads matching information along directed edges in the knowledge graph (KG).
    
    \textbf{DECAF}~\citep{yu2022decaf} arrives at the final answer by co-generating logical forms and direct answers and combining the best of both. Unlike approaches that rely on entity linking tools, DECAF simplifies the process of information retrieval by linearizing the knowledge base into text documents and locating relevant subgraphs using text-based retrieval methods.
    
    \textbf{RoG}~\citep{luo2023reasoning} is an approach that combines LLMs with KG to achieve reliable and interpretable reasoning. The method first generates knowledge graph-based relational paths that serve as faithful reasoning plans, and then utilizes these plans to retrieve valid reasoning paths from the knowledge graph for accurate reasoning in LLMs. RoG enhances the reasoning capabilities of LLMs by training to distill knowledge from knowledge graphs and allows them to be seamlessly integrated with arbitrary LLMs for reasoning.
    
    \textbf{ToG}~\citep{sun2023think} proposes a new LLM-KG integration paradigm “LLM $ \bigotimes $ KG” that treats a LLM as an agent that performs a beam search over the knowledge graph iteratively to discover the most promising reasoning paths and return the most possible reasoning results. ToG leverages the reasoning power of LLMs and expert feedback to ensure traceability and correctability of knowledge. The framework is flexible and plug-and-play for different LLMs, knowledge graphs, and cueing strategies with no additional training cost.
\end{itemize}

\section{Prompts}\label{prompt}
Our proposed KARPA framework consists of the following three main steps: (1) Pre-Planning; (2) Matching; (3) Reasoning.
Among them, steps (1) and (3) use the Large Language Model (LLM), and Appendix~\ref{prompt} provides the related Prompts.

\subsection{Pre-Planning}

\subsubsection{Initial-Planning Prompt}\label{inni_plan}

In the pre-planning stage, initial planning involves using an LLM to preliminarily generate several relation paths of different lengths. The prompt used for this process is given in Content~\ref{inni_plan}.

\newtcolorbox[auto counter, number within=section]{inniplanbox}[2][]{%
  colback=white, 
  colframe=black, 
  width=\linewidth,
  arc=2mm, 
  boxrule=0.5mm, 
  title={\normalsize\faMapSigns\hspace{0.5em}#2},
  breakable, 
  fonttitle=\bfseries\small, 
  fontupper=\footnotesize, 
  #1
}

\begin{inniplanbox}{Initial-Planning Prompt}
   
\hspace{1em}In the process of answer retrieval using a knowledge graph, please think step-by-step and generate reasoning paths of lengths 1, 2, and 3 from a given question and the provided head entity (or entities) that could potentially lead to answer entities. If a reasoning path of the specified length does not exist, please explain the reason.

\vspace{1em}

\textbf{Q}: 

\hspace{1em}Name the president of the country whose main spoken language was Brahui in 1980?

\hspace{1em}Topic Entity: Brahui Language

\textbf{A}:

\hspace{1em}Length 1 reasoning path: The answer entity cannot be reached within a single step, so the length 1 reasoning path is None: \{\}.

\hspace{1em}Length 2 reasoning path: The answer entity may be reached by first finding the corresponding country through the relation "language.human language.main country", and then finding the president of the country through the relation "government.government position held.office holder". So the length 2 reasoning path is: \{language.human\_language.main\_country, government.government\_position\_held.office\_holder\}.

\hspace{1em}Length 3 reasoning path: The answer entity does not require 3 steps to reach, so the length 3 reasoning path is None: \{\}.

\vspace{1em}

\textbf{Q}: 

\hspace{1em}Who is Tom's wife?

\hspace{1em}Topic Entity: Tom

\textbf{A}:

\hspace{1em}Length 1 reasoning path: The answer entity can be reached within a single step by finding Tom's spouse through the relation "people.person.spouse\_s". Therefore, the length 1 reasoning path is: \{people.person.spouse\_s\}.

\hspace{1em}Length 2 reasoning path: The answer entity of the question may be reached if we first find the children through first relation "people.person.children", and then find the parent through second relation "people.person.parent". Therefore, the length 2 reasoning path is: \{people.person.children, people.person.parent\}.

\hspace{1em}Length 3 reasoning path: The answer entity of the question does not require 3 steps to reach, so the length 3 reasoning path is None: \{\}.

\vspace{1em}

\textbf{Q}:

\hspace{1em}\textit{\{A Question.\}}

\hspace{1em}Topic Entity: \textit{\{An Entity\}}

\textbf{A}:

\end{inniplanbox}

\subsubsection{Re-Planning Prompt}\label{re_plan}

In the re-planning  of pre-planning, the LLM is used to re-plan relation paths based on the extracted relations (specifically the top-$K$ relations), which are then used as retrieval information in the matching step. The prompt used is shown in Content~\ref{re_plan}.

\newtcolorbox[auto counter, number within=section]{replanbox}[2][]{%
  colback=white, 
  colframe=black, 
  width=\linewidth,
  arc=2mm, 
  boxrule=0.5mm, 
  title={\normalsize\faMap\hspace{0.5em}#2},
  breakable, 
  fonttitle=\bfseries\small, 
  fontupper=\footnotesize,
  #1
}

\begin{replanbox}{Re-Planning Prompt}
   
\hspace{1em}Given a set of relations and a question, please select relevant relations from the provided relation set to form reasoning paths of length 1, 2, and 3 that could lead from the provided topic entity (or entities) to potential answer entities in a knowledge graph. Ensure that the reasoning paths you create logically connect the topic entity and potential answer entities. Only consider length 3 paths if shorter paths are insufficient to reach the answer. If a reasoning path of the specific length cannot be formed, please explain why.

\vspace{1em}

\textbf{Q}:  

\hspace{1em}Name the president of the country whose main spoken language was Brahui in 1980?

\hspace{1em}Topic Entity: Brahui Language

\hspace{1em}Relations: 

\hspace{2em}language.human\_language.language\_family; 

\hspace{2em}language.human\_language.main\_country; 

\hspace{2em}base.rosetta.languoid.parent; 

\hspace{2em}language.human\_language.writing\_system; 

\hspace{2em}language.human\_language.countries\_spoken\_in; 

\hspace{2em}kg.object\_profile.prominent\_type; 

\textbf{A}:  

\hspace{1em}Length 1 reasoning path: The provided relations cannot reach the answer entity in one step, so the length 1 reasoning path is None: \{\}.

\hspace{1em}Length 2 reasoning path: The answer entity may be reached by first finding the corresponding country through the provided relation "language.human language.main country", and then finding the president of the country through the relation "government.government position held.office holder". So the length 2 reasoning path is: \{language.human\_language.main\_country, government.government\_position\_held.office\_holder\}.

\hspace{1em}Length 3 reasoning path: The answer entity does not require 3 steps to reach, so the length 3 reasoning path is None: \{\}.

\vspace{1em}

\textbf{Q}:  

\hspace{1em}Who is Tom's wife?

\hspace{1em}Topic Entity: Tom

\hspace{1em}Relations: 

\hspace{2em}people.person.profession; 

\hspace{2em}people.marriage.spouse; 

\hspace{2em}people.person.nationality; 

\hspace{2em}award.award\_nomination.award\_nominee; 

\hspace{2em}people.person.parents; 

\hspace{2em}award.award\_nominee.award\_nominations; 

\hspace{2em}people.person.children;

\textbf{A}:  

\hspace{1em}Length 1 reasoning path: Tom's wife in knowledge graph could be reached within a single step by finding Tom's spouse through the provided relation "people.person.spouse\_s". Therefore, the length 1 reasoning path is: \{people.person.spouse\_s\}.

\hspace{1em}Length 2 reasoning path: Tom's wife may be reached if we first find the children through the relation "people.person.children", and then find the parent through second relation "people.person.parent". Therefore, the length 2 reasoning path is: \{people.person.children, people.person.parent\}.

\hspace{1em}Length 3 reasoning path: The answer entity of the question does not require 3 steps to reach, so the length 3 reasoning path is None: \{\}.

\vspace{1em}

\textbf{Q}:

\hspace{1em}\textit{\{A Question.\}}

\hspace{1em}Topic Entity: \textit{\{An Entity.\}}

\hspace{1em}Relations: \textit{\{A list of Relations.\}}

\textbf{A}: 

\end{replanbox}

\subsection{Reasoning}\label{reason_plan}

In the reasoning step, the top-$K$ relation paths retrieved in the matching step, along with their connected topic entity, answer entities, the corresponding question, and all related information are input into the LLM. The prompt used is provided in content~\ref{reason_plan} below.

\newtcolorbox[auto counter, number within=section]{reasonbox}[2][]{%
  colback=white, 
  colframe=black, 
  width=\linewidth,
  arc=2mm, 
  boxrule=0.5mm, 
  title={\normalsize\faPuzzlePiece\hspace{0.5em}#2},
  breakable, 
  fonttitle=\bfseries\small, 
  fontupper=\footnotesize,
  #1
}

\begin{reasonbox}{Reasoning Prompt}
   
\hspace{1em}Given a question and the associated retrieved knowledge graph reasoning paths (topic entity, relation path, tail entity/entities), please think step-by-step and determine whether the tail entity/entities of each provided reasoning paths are the right answer to the question. If any of the tail entities are the correct answer, return those tail entities in the format of {answer\_1, answer\_2, ...}. If the tail entity of current reasoning path is not the answer of question, explain why. All reasoning paths should be analysed.

\vspace{1em}

\textbf{Q}:  

\hspace{1em}Rift Valley Province is located in a nation that uses which form of currency?

\hspace{1em}Reasoning Paths:

\hspace{2em}(Rift Valley Province, location.administrative division.country, Kenya)

\hspace{2em}(Rift Valley Province, location.location.geo location, UnName Entity)

\hspace{2em}(Rift Valley Province, location.administrative division.country $ \rightarrow $ location.country.currency used, Kenyan shilling)

\textbf{A}: 

\hspace{1em}Let's analyze the reasoning paths step-by-step to determine the correct answer to the question.

\hspace{1em}1.(Rift Valley Province, location.administrative division.country, Kenya): Rift Valley Province is located in Kenya. However, it does not provide information about the currency used in Kenya, so Kenya is not the direct answer to the question.

\hspace{1em}2.(Rift Valley Province, location.location.geo location, UnName Entity): The UnName Entity is not the answer of question.

\hspace{1em}3.(Rift Valley Province, location.administrative division.country$ \rightarrow $ location.country.currency used,

Kenyan shilling): Kenyan shilling is the form of currency used in the nation where Rift Valley Province is located, so 'Kenyan shilling' is the right answer to the question.

\hspace{1em}The correct answer to the question is the Kenyan shilling, as identified in the fourth reasoning path. Therefore, the correct tail entity is:

\{Kenyan shilling\}.

\vspace{1em}

\textbf{Q}:   

\hspace{1em}\textit{\{A Question.\}}

\hspace{1em}Reasoning Paths: \textit{\{A list of Reasoning Paths.\}}

\textbf{A}:

\end{reasonbox}

\begin{figure*}[h]
\centering
\includegraphics[width=0.9\textwidth]{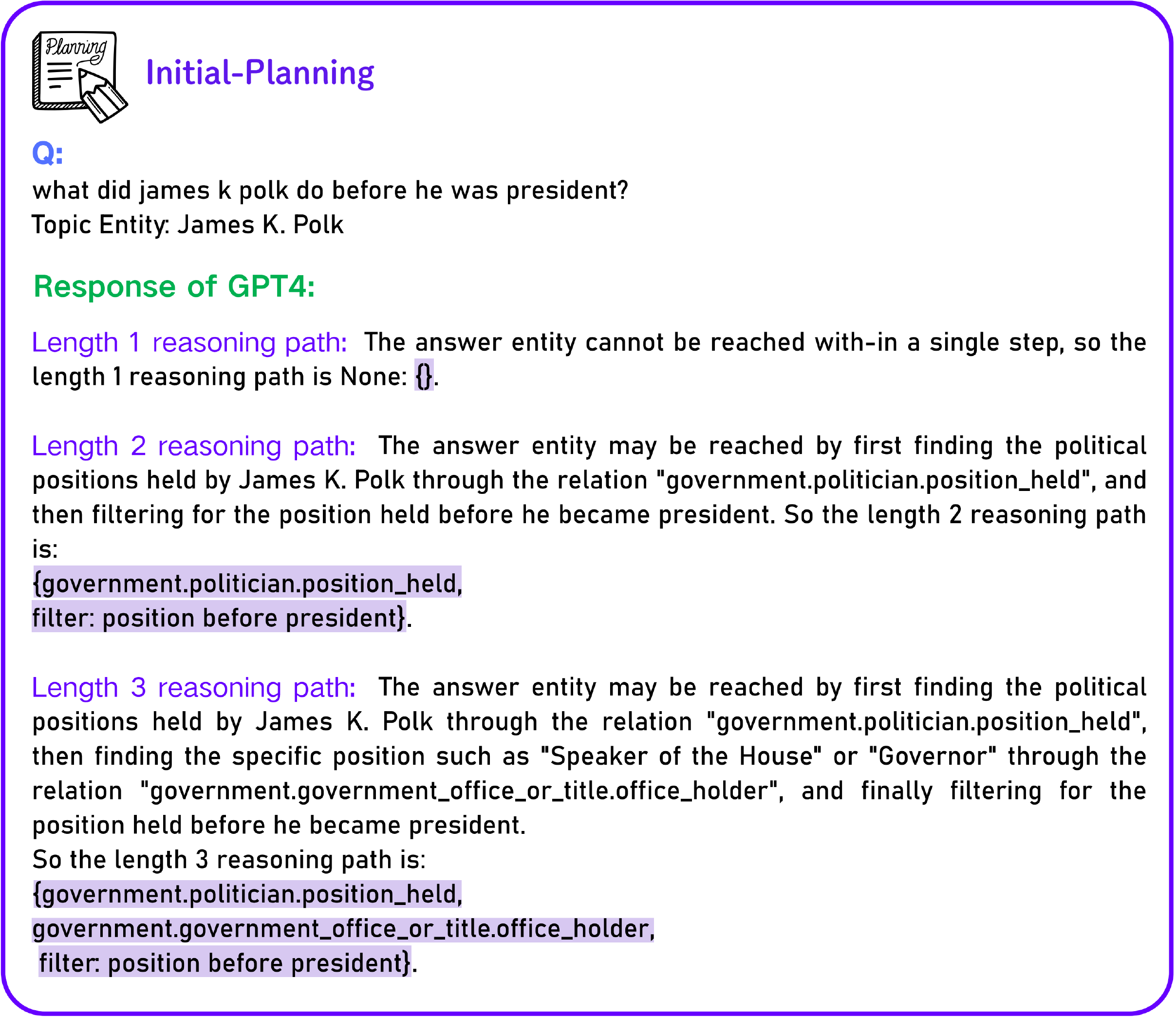}
\caption{A Case for Relational Path Generation in the Initial-Planning Phase.}
\label{case_1}
\end{figure*}

\begin{figure*}[h]
\centering
\includegraphics[width=0.87\textwidth]{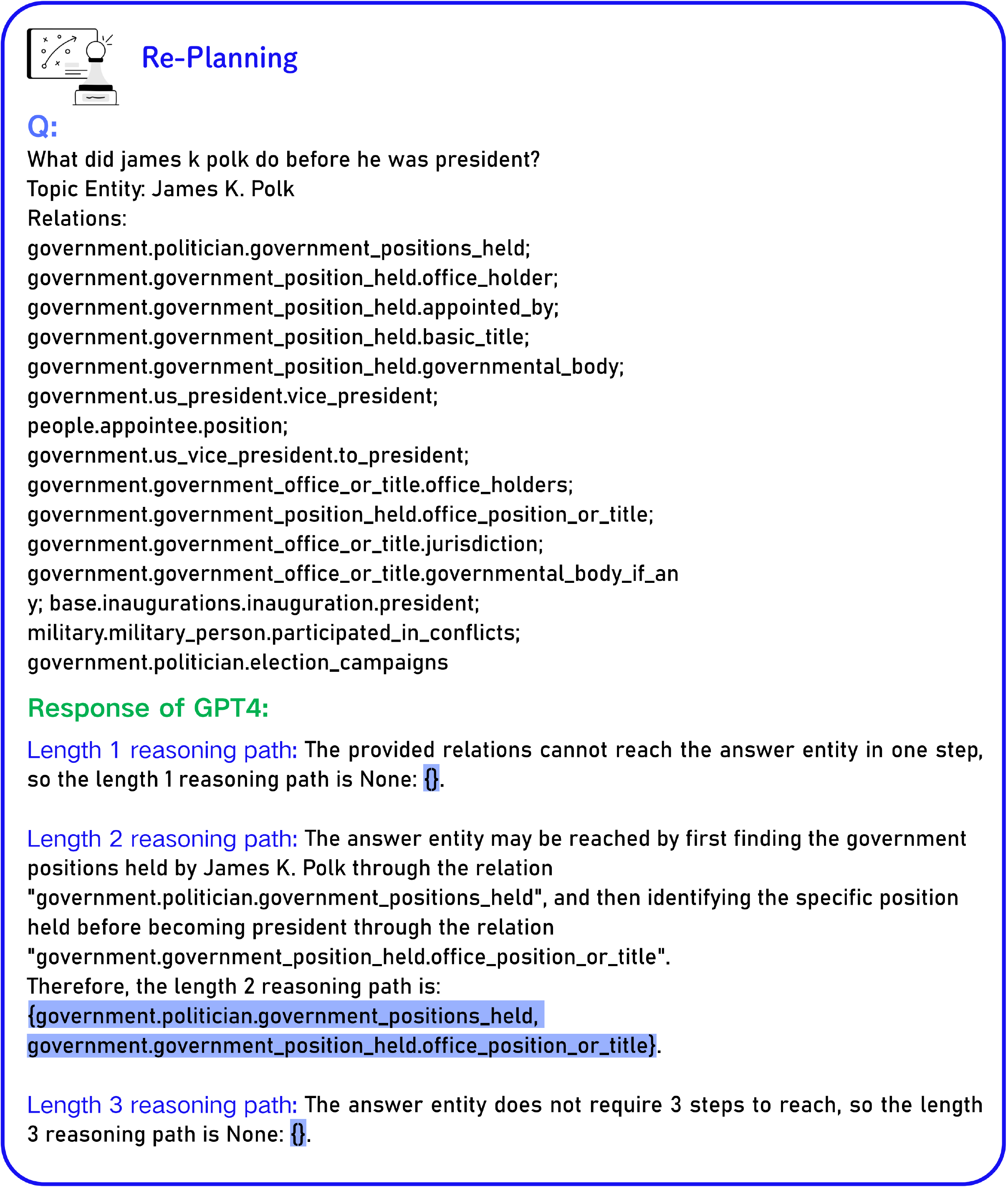}
\caption{A Case of Relation Path Assembling in the Re-Planning Phase.}
\label{case_2}
\end{figure*}

\begin{figure*}[h]
\centering
\includegraphics[width=0.9\textwidth]{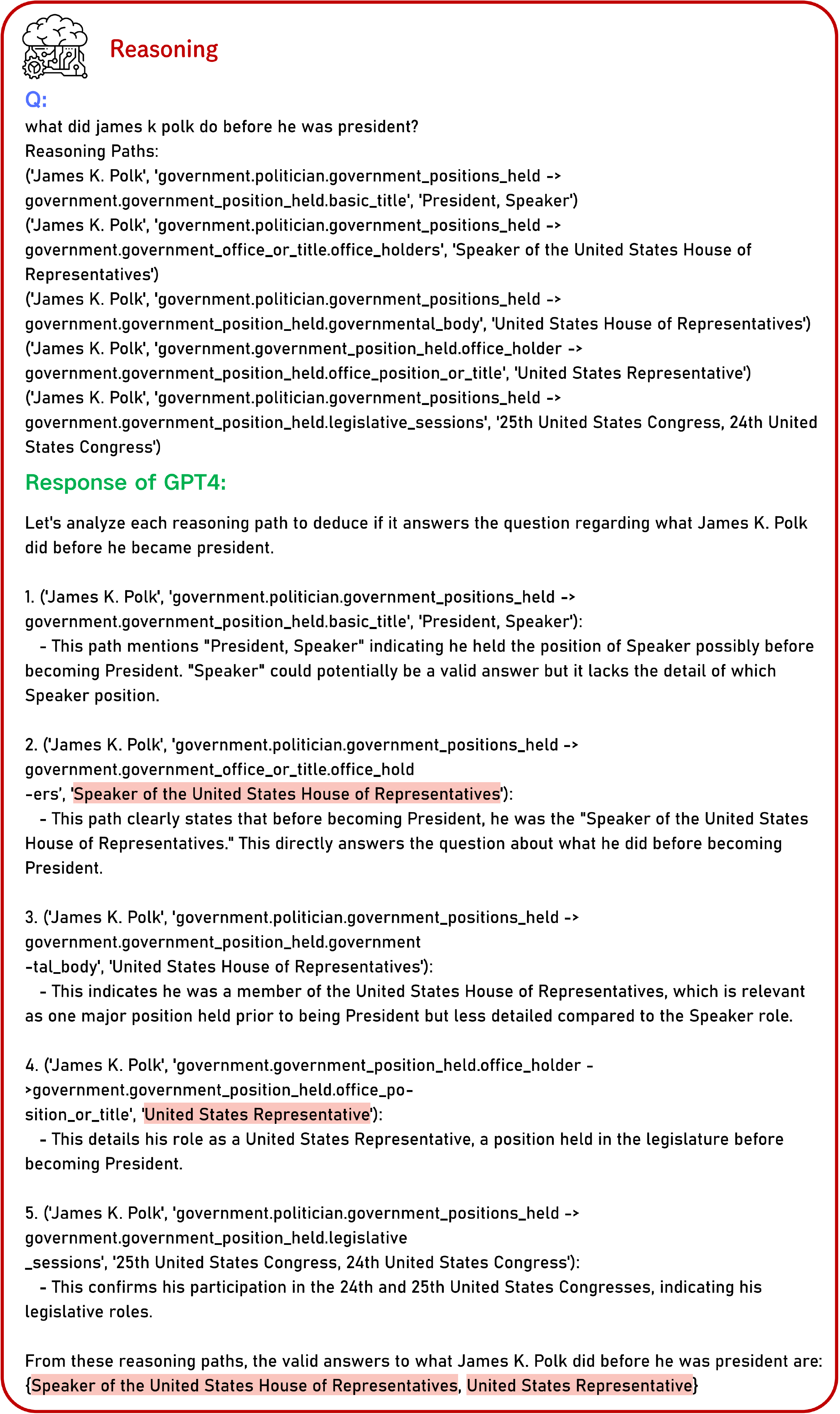}
\caption{A Case of Inputting Top-$K$ Relation Paths into LLMs During Reasoning Phase.}
\label{case_3}
\end{figure*}

\section{Case Study}

In this section, we present a detailed case study to illustrate the effectiveness of KARPA in handling complex knowledge graph question answering (KGQA) tasks. KARPA leverages LLMs in both the pre-planning and reasoning steps. For the question ``What did James K. Polk do before he was president?", KARPA uses the LLM to generate initial reasoning paths and then further refines the answer by reasoning over the identified relation paths and corresponding entities. The following case study elaborates on the workflow of KARPA in this example, showcasing its ability to utilize external knowledge and LLM planning capabilities to accurately answer the question.

In the pre-planning step, KARPA first utilizes the LLM to generate initial relational paths based on the provided question, as shown in Figure~\ref{case_1}. Given the question ``What did James K. Polk do before he was president?", the LLM generates paths of varying lengths. Initially, the LLM considers whether the answer entities can be reached within a single relational step. Since the LLM considers the answer entities for this question cannot be reached in one step, the LLM outputs an empty reasoning path of length 1. 

When considering a relational path with two associated relations, the LLM infers that the answer entity can be found by first identifying the political positions held by James K. Polk through the relation ``government.politician.position\_held," and then filtering for the position he held before becoming president using ``filter: position before president." Thus, the LLM determines that the answer entities can be reached via the path \{government.politician.position\_held, filter: position before president\}. Additionally, the LLM considers that the answer entities might be accessible through a path involving three relations. This step-by-step reasoning process allows the LLM to initially plan multiple reasoning chains for subsequent relation extraction.

In the third phase of the pre-planning step, KARPA employs the LLM to re-plan the relational paths based on the set of extracted relations. For the question ``What did James K. Polk do before he was president?", the LLM is provided with a set of relations, as illustrated in Figure~\ref{case_2}. The LLM is tasked with selecting relevant relations from the list and assembling them into complete reasoning chains that potentially connect the topic entity to the answer entities.

In this case, the LLM determines that the answer entities cannot be reached using a single relation from the provided list, and therefore outputs an empty relation path for length 1. When constructing a relation path of length 2, the LLM identifies that ``government.politician.government positions held" and ``government.government position held.office position or title" form a complete reasoning chain, enabling the extraction of the correct answer entities for the given question. As a result, the LLM outputs the length 2 relation path as \{government.politician.government\_positions\_held, government.government\_position\_held.office\_position or\_title\}. Since the LLM considers that the answer can be extracted using this two-step reasoning chain, it determines that a three-step reasoning chain is unnecessary and outputs None for the length 3 relation paths.

In the reasoning step of KARPA, several candidate relational paths are provided for the LLM to determine the final answer. Given these candidate paths and their corresponding entities, the LLM analyzes each path step-by-step, enabling more thoughtful and accurate reasoning. An example of KARPA's reasoning process is illustrated in Figure~\ref{case_3}.

\end{document}